\documentclass{template/cambrian}
\usepackage{template/swam-layout}
\usepackage{placeins}
\usepackage{wrapfig}
\usepackage{multirow}
\usepackage{template/swam-highlights}
\newcommand{\PaperName}{SWAM}
\newcommand{\PaperSubtitle}{Scaling World Action Models with Cross-Modal Alignment}
\newcommand{\PaperAuthors}{Yu Zhou\textsuperscript{1,2,3,*,\textdaggerdbl}, Xinyuan Xia\textsuperscript{1,2,3,*}, Xuankun Yang\textsuperscript{1,3,*}, Yinghao Yang\textsuperscript{4,2,3,*}, Wendi Chen\textsuperscript{1,2,3}\\Bowen Liu\textsuperscript{1,2,3,\textdaggerdbl}, Jianhan Ma\textsuperscript{3}, Weicheng Duan\textsuperscript{5,3}, Jun Lv\textsuperscript{3}, Cewu Lu\textsuperscript{1,2,3}, Yanwei Li\textsuperscript{1,\textdagger}}
\newcommand{\PaperAffiliations}{\textsuperscript{1}Shanghai Jiao Tong University \quad \textsuperscript{2}Shanghai Innovation Institute \quad \textsuperscript{3}Noematrix Ltd.\\\textsuperscript{4}Westlake University \quad \textsuperscript{5}Tongji University \quad \textsuperscript{*}Equal contribution. \quad \textsuperscript{\textdagger}Corresponding author.}
\newcommand{\PaperAuthorNote}{\textsuperscript{\textdaggerdbl}Work completed during internships at Shanghai Jiao Tong University.}
\newcommand{\PaperDate}{November 9, 2026} 
\newcommand{\PaperAbstract}{
World action models jointly predict video and actions conditioned on language. 
Existing studies focus primarily on architectures and training mechanisms, typically treating language--video alignment acquired through video pretraining as an established foundation. 
However, introducing action learning may impair this alignment. 
Principled guidance on data and supervision design for maintaining language--video alignment during action learning remains limited. 
To investigate this, we curate a pool of approximately 47,000 hours of general videos, egocentric videos, UMI demonstrations, and robot demonstrations.
Building on this, we systematically study how language--video supervision, language--video--action supervision, and the joint use of these two forms affect WAM performance. 
In particular, we examine how language annotation quality affects pretrained language--video capabilities, and how action supervision from UMI and robot demonstrations extends these capabilities to language--video--action modeling. 
To better scale the model, we further conduct detailed studies on joint training recipe and model architecture.
Under a fixed training budget, mixing egocentric videos and UMI demonstrations improves control generalization.
Model architecture affects how effectively these experiences are used, while scaling pretraining under the selected configuration brings further gains.
The resulting recipe, post-trained on RoboTwin Clean demonstrations only, achieves success rates of 88.10\% and 61.56\% under Easy (Clean-to-Clean) and Hard (Clean-to-Random) conditions, respectively. Our model achieves 40.00\% SR across three real-world tasks, compared with 23.33\% for the strongest baseline, indicating better generalization under unseen conditions.
We release the curated data, training code, and model weights.
}
\newcommand{\PaperTeaserCaption}{\textbf{Overview of SWAM.} We study how multisource supervision preserves the language--video capabilities acquired during video pretraining and extends them to robot control. We curate a multisource data pool of approximately 47,000 hours to support this study. Experiments show that additional pretraining further improves real-world performance, and that the resulting recipe generalizes to unseen scene conditions in RoboTwin and on real robots.}
\newcommand{\PaperWebsite}{https://char-lotte-xia.github.io/swam-official/}
\newcommand{\PaperCode}{https://github.com/shaaaaron/SWAM}
\newcommand{\PaperData}{https://huggingface.co/datasets/FreDERicia/SWAM-Pretrain}
\newcommand{\TeaserFile}{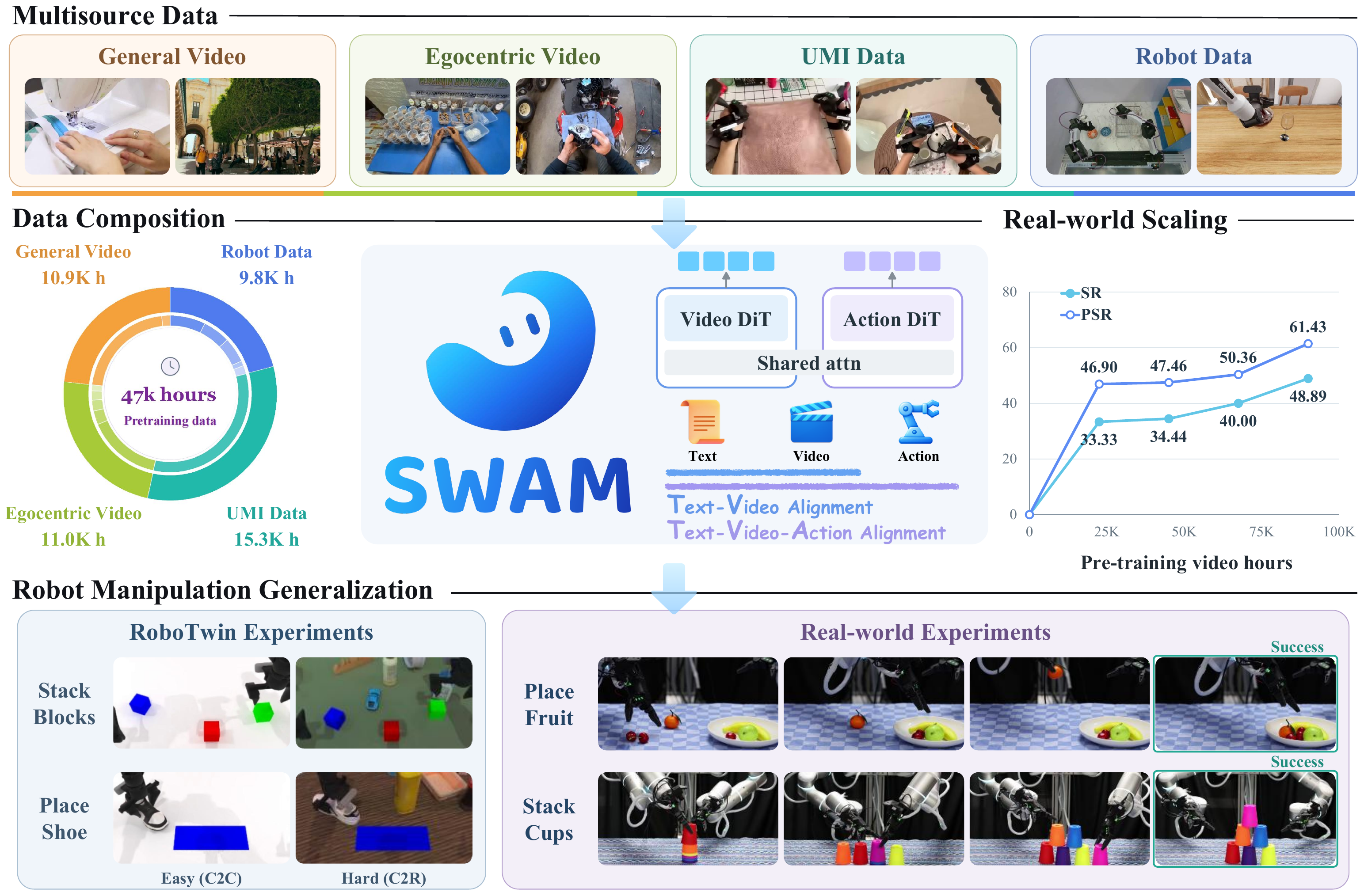}

\begin{document}
\makecover
\restoregeometry
\captionsetup{
  font=small,
  labelsep=period
}
\begingroup
  \hypersetup{linkcolor=black}
  \microtypesetup{protrusion=false}
  \hyphenpenalty=10000\exhyphenpenalty=10000\emergencystretch=2em
  \tableofcontents
\endgroup
\clearpage
\section{Introduction}

Pretrained video models learn associations between language descriptions, visual content, and visual dynamics, acquiring visual and dynamics priors across scenes, objects, and interactions \cite{wu2025hunyuanvideo,wan2025wan,DBLP:conf/iclr/YangTZ00XYHZFYZ25,seedance2026seedance}. These experiences provide a foundation for robots to understand manipulation goals and predict environmental changes \cite{wu2024unleashing,DBLP:conf/icml/HuGWCWZSL025}. World action models (WAMs) jointly predict video and actions conditioned on language, connecting visual changes to robot actions and providing a route from video pretraining to robot control \cite{ye2026world,li2026causal}. This paper studies how to better use these priors to extend language-conditioned video modeling into generalizable action capabilities.

When we introduce action prediction, the language--video capabilities acquired in video pretraining should be preserved. 
This requires further alignment between language, video, and action.
It bridges the gap between language-described targets, visual observation, and executed actions. 
Existing work on WAMs largely focused on model architectures and training mechanisms \cite{yu2026openwam,sun2026ego4wam}. 
Methods on video pretraining often treat the language--video alignment as an established foundation \cite{ye2026world,li2026causal}.
However, introducing action learning raises a basic question: how to preserve established language--video prior and further solve language--video--action misalignment? 
The roles of different experiences and forms of supervision in this process, particularly for control generalization, remain unclear \cite{yu2026openwam,sun2026ego4wam,yeom2026makes,jiang2026rethinking}. 

\begin{figure}[!htbp]
\centering
\includegraphics[width=\linewidth]{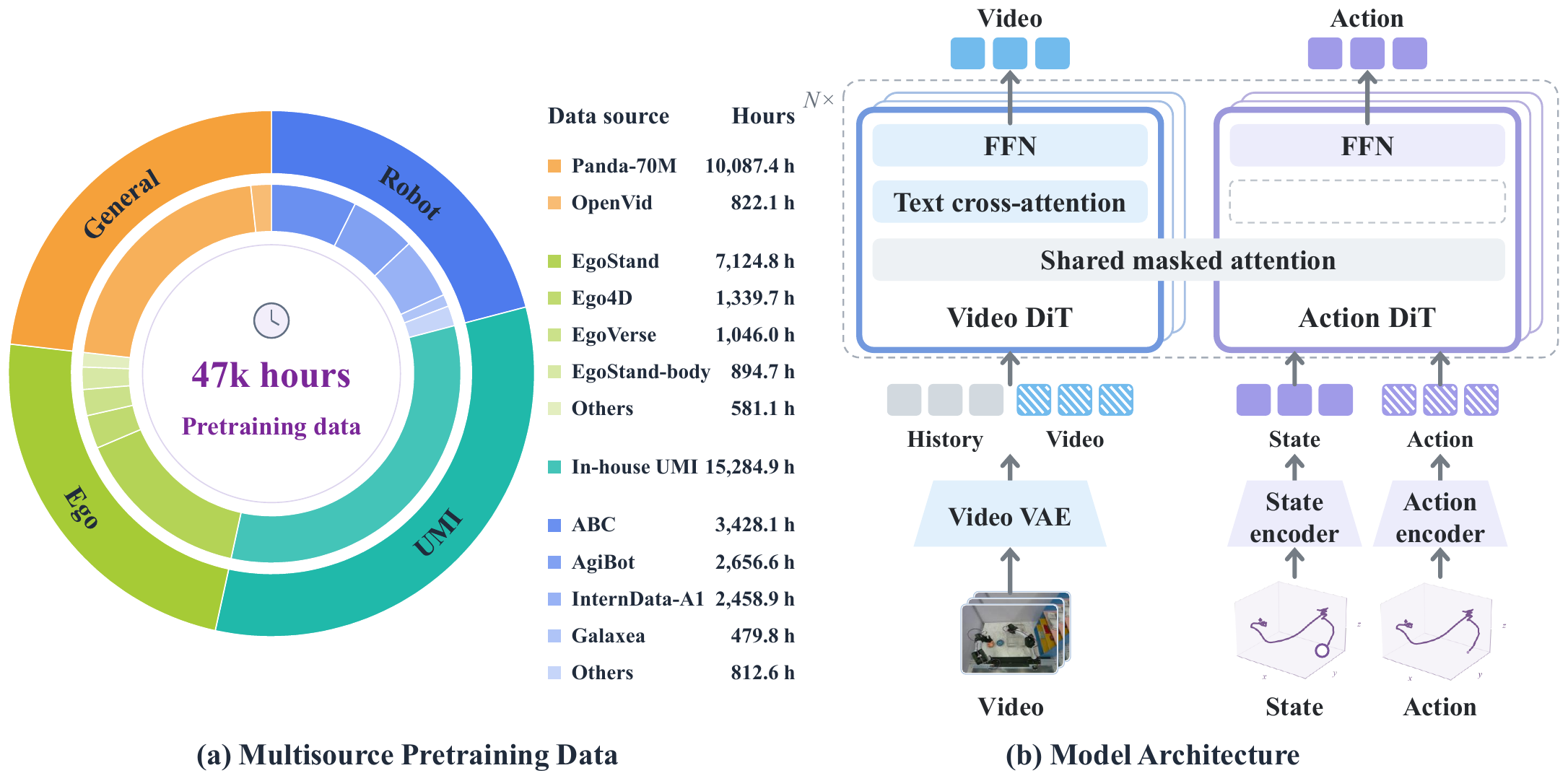}
\caption{\textbf{Multisource data and model overview.} (a) Composition of the 47k-hour pretraining data pool. (b) Our world action model architecture, with video and action branches connected through shared masked attention.}
\label{fig:intro-data-model}
\end{figure}

To study these questions, we first construct a multisource data pool with approximately 47,000 hours, comprising approximately 11,000 hours of general videos, 11,000 hours of egocentric videos, 15,000 hours of in-house UMI demonstrations, and 10,000 hours of robot demonstrations, as shown in Figure~\hyperref[fig:intro-data-model]{\ref*{fig:intro-data-model}(a)}. 
These sources provide different forms of cross-modal supervision. 
General and egocentric videos primarily provide language--video supervision, covering general visual dynamics and human interactions, respectively. 
UMI demonstrations record video and action trajectories of humans manipulating objects with handheld grippers, while robot demonstrations provide observations, states, and actions during robot task execution. 
Both include language descriptions corresponding to the manipulation process, supporting language--video--action learning in addition to language--video supervision. 
UMI demonstrations provide scalable human manipulation experience, whereas robot demonstrations provide direct but more limited robot control experience \cite{DBLP:conf/rss/ChiXPCB0TS24,zhaxizhuoma2025fastumi,fang2026robopocket}. The sources differ in scene and task coverage and in the completeness of language and action supervision. Multisource pretraining therefore requires combining these experiences appropriately: language--video supervision preserves video pretraining capabilities, while human and robot action supervision establishes video--action correspondences \cite{wei2026xpace,miao2026internw0,guo2026video}.

Building on the constructed multisource experience, we conduct systematic studies at three levels: language--video alignment, language--video--action alignment, and joint learning. 
Specifically, at the language--video level, we examine how to preserve the visual prior in pretrained video generation with the design of language supervision. 
At the language--video--action level, we investigate how to utilize action supervision from UMI and robot demonstrations to help action prediction and robot control. 
In joint training, we study the data recipe and model architecture for better cross-modality modeling. 
The data recipe determines the coverage of the visual and interaction experience in tasks, scenes, and the proportions of different experiences.
The model architecture answers how to conduct cross-modal supervision through modality interactions and language conditioning pathways, as shown in Figure~\hyperref[fig:intro-data-model]{\ref*{fig:intro-data-model}(b)}. 
We further examine the benefits of scaling training with the resulting data and supervision recipe.

Our study shows that preserving language--video capabilities and extending language--video--action capabilities jointly support control generalization in WAMs. 
Detailed manipulation descriptions and environmental cues improve language--video supervision, which maintains language--video alignment and thereby improves control generalization. 
And UMI demonstrations improve robot control in simulation and real scenes, proving the value of human manipulation experience with action supervision for robot learning. 
In joint learning, the benefits of multisource experience depend both on data and model  supervision recipe. 
Experiments show that increasing training investment under the selected data and model recipe further improves control generalization. 
These findings provide empirical guidance for data collection and organization, supervision design, and joint training in WAMs. The pretraining recipe derived from these findings achieves success rates of 88.10\% and 61.56\% under the RoboTwin Easy (Clean-to-Clean) and Hard (Clean-to-Random) settings, respectively. Ours achieves 40.00\% SR under the Hard setting across three real-world tasks, exceeding OpenWAM's 23.33\%. This result indicates better generalization to unseen objects and backgrounds.



\begingroup
\setlength{\emergencystretch}{2em}
\section{Related Work}

\noindent\textbf{World Action Models.} World action models aim to connect visual observations with executable robot actions \cite{ye2026world}. Early work treated video as a transferable visual plan or dynamic representation. UniPi first generates a language-conditioned visual plan and then recovers actions with an inverse-dynamics module \cite{DBLP:conf/nips/DuY0DN0SA23}. VPP conditions a policy on predictive representations from a video model \cite{DBLP:conf/icml/HuGWCWZSL025}. Motus, DreamZero, and LingBot-VA place visual dynamics and action prediction within unified or interactive architectures, exploring discrete tokens, latent actions, joint flow matching, and causal video--action modeling \cite{bi2026motus,ye2026world,li2026causal}. Fast-WAM and InternW0-$\Delta$ further distinguish future supervision during training from action generation at inference, showing that the prediction target and the deployment interface can be separated \cite{yuan2026fast,miao2026internw0}. VPP2 and LingBot-VA adapt video pretraining to robot manipulation by studying how manipulation data organization, temporal and video representation modeling, and action-branch adaptation can preserve video prediction capabilities while supporting action learning \cite{guo2026video,li2026causal}. We build on this foundation and focus on how language--video capabilities from video pretraining can transfer to language--video--action learning and action control.

\noindent\textbf{Multisource Data for Robot Learning.} Robot learning increasingly draws on cross-embodiment trajectories, general videos, egocentric videos, and human manipulation demonstrations rather than a single source of robot data. Open X-Embodiment, Octo, and OpenVLA aggregate and filter multi-robot data to study shared representations, observation adaptation, and action-space alignment, showing that cross-embodiment learning requires source normalization and interface alignment \cite{collaboration2023open,DBLP:conf/rss/GhoshWPBMDHK0LT24,kim2024openvla}. General and egocentric videos expand coverage of scenes, objects, and interaction dynamics, but usually lack action labels directly corresponding to robot execution \cite{grauman2022ego4d,punamiya2026egoverse}. EgoMimic and EgoScale study how such human experience can be transferred through view and embodiment alignment, including settings with limited robot data \cite{kareer2025egomimic,zheng2026egoscale}. UMI complements these sources by using portable interfaces to record manipulation trajectories in an end-effector-like interaction format, thereby providing action supervision associated with the manipulation process \cite{DBLP:conf/rss/ChiXPCB0TS24,zhaxizhuoma2025fastumi,fang2026robopocket}. EgoWAM, Ego4WAM, and Motus further demonstrate the complementary roles of video, action trajectories, semantic descriptions, and world-prediction targets \cite{li2026egowam,sun2026ego4wam,bi2026motus}. The central issue in multisource learning is therefore not simply adding sources, but understanding differences in scene coverage, interaction dynamics, embodiment interfaces, and available supervision. We curate approximately 47,000 hours of multisource data. We use this pool to systematically examine how experience combinations and supervision design affect WAM pretraining.


\endgroup

\section{Method}

This section presents the world action model and its multimodal supervision. We first present the problem formulation, model architecture, and training objectives, then describe multisource experience curation and supervision construction. Pretraining uses four types of experience: general videos, egocentric videos, UMI demonstrations, and robot demonstrations. These sources differ in scene and task distributions, data quality, manipulation diversity, and available action supervision. General and egocentric videos provide language--video supervision, while UMI and robot demonstrations additionally provide state and action signals where available and valid. In Table~\ref{tab:pretraining-inventory} and Figure~\ref{fig:pretraining-composition}, we summarize the composition, scale, and video-duration distribution of the pretraining pool after data curation and cleaning.

\begin{table}[!htb]
\raggedright
\begin{minipage}[t]{0.49\linewidth}
\vspace{0pt}
\centering
\captionsetup{labelsep=period}
\caption{\textbf{Composition and scale of multisource pretraining data.}}
\label{tab:pretraining-inventory}
\footnotesize
\setlength{\tabcolsep}{3pt}
\renewcommand{\arraystretch}{1.12}
\begin{tabular*}{\linewidth}{@{\extracolsep{\fill}\hspace{\tabcolsep}}lrrr}
\toprule[0.8pt]
\textbf{Dataset} & \textbf{Episodes} & \textbf{Frames (M)} & \textbf{Hours} \\
\midrule[0.5pt]
\multicolumn{3}{l}{\textit{General Videos}} & \textbf{10,909.6} \\
\hspace{0.8em}Panda-70M & 2,599,835 & 1,089.4 & 10,087.4 \\
\hspace{0.8em}OpenVid & 256,839 & 82.5 & 822.1 \\
\midrule[0.4pt]
\multicolumn{3}{l}{\textit{Egocentric Videos}} & \textbf{10,986.3} \\
\hspace{0.8em}EgoSuite & 294,798 & 928.9 & 8,600.6 \\
\hspace{0.8em}Ego4D & 21,302 & 144.7 & 1,339.7 \\
\hspace{0.8em}EgoVerse & 55,112 & 113.0 & 1,046.0 \\
\midrule[0.4pt]
\multicolumn{3}{l}{\textit{UMI Data}} & \textbf{15,284.9} \\
\hspace{0.8em}In-house UMI & 411,754 & 1,650.8 & 15,284.9 \\
\midrule[0.4pt]
\multicolumn{3}{l}{\textit{Robot Data}} & \textbf{9,836.0} \\
\hspace{0.8em}ABC & 127,057 & 370.2 & 3,428.1 \\
\hspace{0.8em}AgiBot & 155,650 & 286.9 & 2,656.6 \\
\hspace{0.8em}InternData-A1 & 354,192 & 265.6 & 2,458.9 \\
\hspace{0.8em}Galaxea & 19,636 & 51.8 & 479.8 \\
\hspace{0.8em}RoboCOIN & 50,439 & 45.3 & 419.1 \\
\hspace{0.8em}RoboMind & 61,382 & 42.5 & 393.5 \\
\midrule[0.4pt]
\textbf{Total} & \textbf{4,407,996} & \textbf{5,071.5} & \textbf{47,016.7} \\
\bottomrule[0.8pt]
\end{tabular*}

\end{minipage}\hfill
\begin{minipage}[t]{0.49\linewidth}
\vspace{0pt}
\centering
\includegraphics[width=\linewidth]{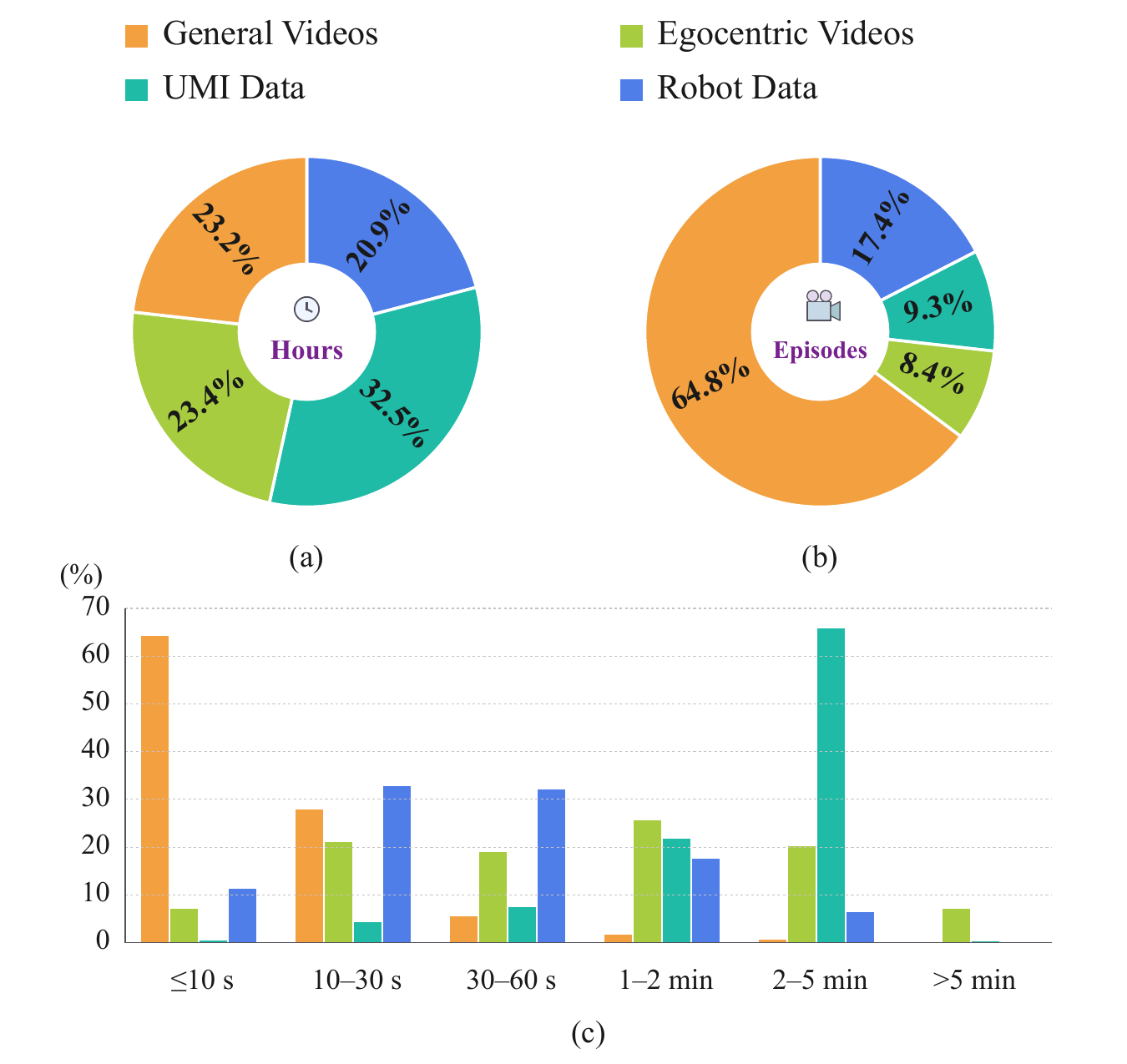}\par
\captionsetup{type=figure,labelsep=period,skip=3pt}
\caption{\textbf{Pretraining data composition and video duration across sources.} (a) Shares of total video hours. (b) Shares of total episodes. (c) Video-duration distribution within each source.}
\label{fig:pretraining-composition}
\end{minipage}
\end{table}

\subsection{World Action Modeling}

\subsubsection{Problem Formulation}

We consider language-conditioned robot manipulation. Given the visual observations acquired so far, a language instruction, and its current proprioceptive state, the robot predicts actions over a future interval. Let $n$ be a prediction start point sampled from a trajectory, $K>0$ a fixed prediction horizon, $V_{0:n}$ the visual history including the current observation, $P$ the language condition, and $S_n$ the current state. The control task requires learning an action policy $\pi_\theta$ conditioned on visual history, the language instruction, and the current state:

\begin{equation}
A_{n:n+K}\sim\pi_\theta(\cdot\mid V_{0:n},P,S_n),
\end{equation}

\noindent where $A_{n:n+K}$ denotes the action sequence over the target interval, and $\theta$ denotes the model parameters.

When trained primarily on observation–action pairs, a control policy must jointly address scene understanding, interaction dynamics, and action generation. We adopt a world action model (WAM) based on a pretrained video model, using the associations between language descriptions and visual dynamics learned during video pretraining and learning correspondences between manipulation processes and executed actions through action supervision. The model jointly predicts video changes and actions over the same interval. The video sequence $V_{n:n+K}$ describes observable scene evolution during manipulation, while the action sequence $A_{n:n+K}$ describes the corresponding manipulation behavior. The two sequences span the same physical time interval but may use different sampling rates.

We denote the joint predictive distribution of video and actions by $p_\theta$:

\begin{equation}
\begin{aligned}
&p_\theta(V_{n:n+K},A_{n:n+K}\mid V_{0:n},P,S_n)\\
&=p_\theta(V_{n:n+K}\mid V_{0:n},P,S_n)\,p_\theta(A_{n:n+K}\mid V_{0:n+K},P,S_n)\\
&=p_\theta(V_{n:n+K}\mid V_{0:n},P)\,p_\theta(A_{n:n+K}\mid V_{0:n+K},P,S_n).
\end{aligned}
\end{equation}

The final equality follows from our modeling choice that the video branch is not conditioned on the proprioceptive state. The first term learns language-conditioned video prediction, corresponding to language--video learning. The second term predicts actions from language, video information, and the current state. Together, the two terms constitute joint language--video--action learning.

Starting from a pretrained video model, we train a unified WAM end to end with a joint prediction objective. The action branch uses visual dynamics and language conditioning information supplied by the video branch. Training retains the video prediction objective while introducing action supervision, aiming to continue using existing language--video modeling capabilities and to learn correspondences between manipulation processes in video and executed actions.

Multisource experience supplies different parts of the supervision for joint prediction. General and egocentric videos provide language--video correspondences for learning language-conditioned visual dynamics. UMI and robot demonstrations additionally provide actions temporally associated with video, enabling action prediction alongside video learning. All samples participate in video learning. Samples with valid action annotations additionally participate in action learning. Missing or invalid action targets contribute no action supervision. Experiences with different supervision coverage can therefore participate in unified WAM pretraining.

\subsubsection{Model Architecture}

\begin{figure}[!htbp]
\centering
\includegraphics[width=0.89\linewidth]{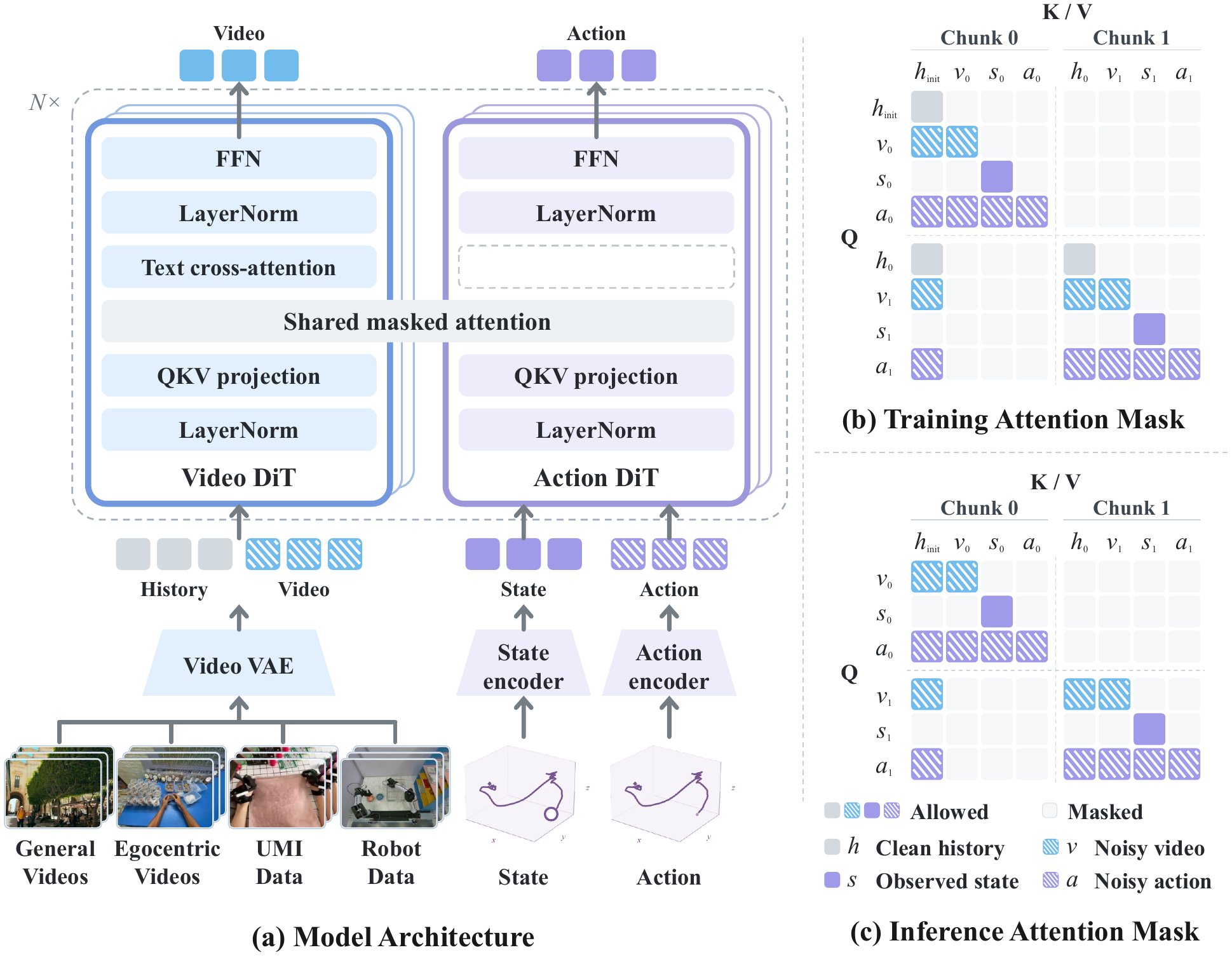}
\caption{\textbf{World action model architecture and attention masks.} (a) The video and action DiT branches are connected through shared masked attention, with text cross-attention in the video branch only. Attention masks for (b) training with clean video history under teacher forcing and (c) generating video and actions chunk by chunk at inference.}
\label{fig:model-architecture}
\end{figure}

Given visual observations $V=\{v_i\}_{i=1}^{T}$, a language instruction $P$, and available states $S$, we adopt a WAM that jointly predicts video and actions. Whereas $n:n+K$ above denotes a trajectory-level prediction interval, we use $t$ below to index temporally aligned video--action chunks. Each chunk contains 2 video latent frames and 48 action steps. For chunk $t$, let $V_t\in\mathbb{R}^{T_v\times H\times W\times 3}$ denote the corresponding video, $S_t\in\mathbb{R}^{D_s}$ the available state at the chunk start, and $A_t\in\mathbb{R}^{H_a\times D_a}$ the normalized target action sequence. Here, $T_v$ is the number of video frames, $H$ and $W$ are frame height and width, $D_s$ is the state dimension, and $H_a$ and $D_a$ are the number of action steps and the action representation dimension, respectively. A VAE encodes video into latents $x_t$. Separate encoders map actions and states to tokens. A text encoder maps the language instruction to condition $p$.

We illustrate the model architecture and the attention masks used during training and inference in Figure~\ref{fig:model-architecture}. The model uses a MoT architecture~\cite{DBLP:journals/tmlr/LiangYL0DZGLYZL25}, with the video branch initialized from Wan \cite{wan2025wan} and the action branch randomly initialized. The action DiT branch comprises action and state encoders, transformer blocks, and an action decoder. Video and action maintain separate residual streams, with joint attention passing information unidirectionally from video to action. The video token stream passes through the video branch and causally attends to current and past video tokens. The action token stream passes through the action branch and attends to video tokens from past and current chunks, as well as state and action tokens within the current chunk. This unidirectional interaction allows samples with video supervision alone to continue training the video branch without directly updating the action branch through the video objective, supporting the use of videos without action supervision.

The video branch receives encoded language conditions through cross-attention. The action branch uses language information only through attention interactions with the video branch and has no direct text cross-attention. This conditioning pathway aims to reuse the language--video correspondences established during video pretraining, allowing action prediction to use semantic information associated with the manipulation process. The video branch predicts flow velocities in video latent space, while the action branch decodes its output tokens into flow velocities in action space.

\subsubsection{Training Objectives}

For chunk $t$, $x_t$ and $A_t$ denote the clean video latents and action targets, respectively. Let $\tau\in[0,1]$ denote flow matching time. We sample independent standard Gaussian noise $\epsilon_v$ and $\epsilon_a$ with the same shapes as their respective targets and construct noisy targets:

\begin{equation}
\begin{aligned}
x_t^{\tau} &= (1-\tau)x_t + \tau\epsilon_v,\\
A_t^{\tau} &= (1-\tau)A_t + \tau\epsilon_a.
\end{aligned}
\end{equation}

The action encoder maps noisy actions $A_t^{\tau}$ to action tokens. Video and actions within a chunk share $\tau$, sampled independently across chunks. Training uses teacher forcing, with past clean video latents $h_t$ as context. Let $u_{\theta}^{v}$ and $u_{\theta}^{a}$ denote the video and action velocity predictors, whose outputs have the same shapes as $x_t$ and $A_t$, respectively. The action predictor $u_{\theta}^{a}$ denotes the complete action DiT branch, including action and state encoding, transformer processing, and action decoding. The two flow matching losses~\cite{DBLP:conf/iclr/LipmanCBNL23} are:

\begin{equation}
\begin{aligned}
\mathcal{L}_{\mathrm{video}} &= \mathbb{E}\!\left[
\left\|u_{\theta}^{v}(x_t^{\tau},h_t,p,\tau)
- (\epsilon_v-x_t)\right\|_2^2\right],\\
\mathcal{L}_{\mathrm{action}} &= \mathbb{E}\!\left[
\left\|u_{\theta}^{a}(A_t^{\tau},x_t^{\tau},h_t,S_t,p,\tau)
- (\epsilon_a-A_t)\right\|_2^2\right].
\end{aligned}
\end{equation}

\noindent Here, $\theta$ denotes model parameters, and the expectations are taken over training samples, noise, and flow matching times. The action output depends on the current video $x_t^{\tau}$ and language condition $p$ through the video branch. The two losses are combined with equal weights:

\begin{equation}
\mathcal{L} = \mathcal{L}_{\mathrm{video}} + \mathcal{L}_{\mathrm{action}}.
\end{equation}

All samples contribute to the video objective. The action objective is computed only for samples with valid action annotations.

\Needspace{5\baselineskip}
\subsection{Language--Video Alignment}
\label{sec:method:lv}

\subsubsection{Video Sources and Curation}
\label{sec:method:video-curation}

\begin{figure}[!htbp]
\centering
\includegraphics[width=\linewidth]{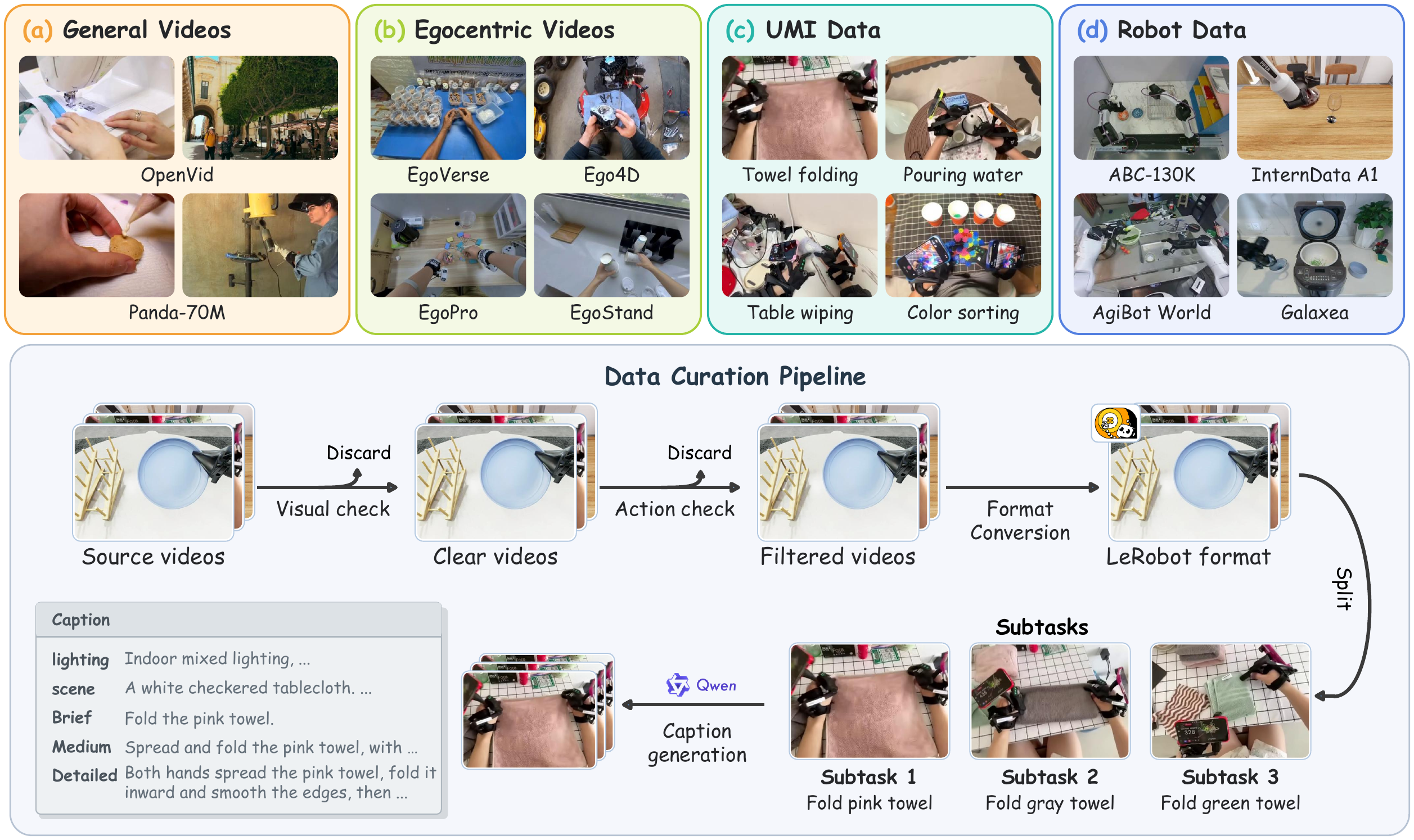}
\caption{\textbf{Multisource training data and curation pipeline.} The upper panel shows representative examples from four data sources. The lower panel illustrates visual checks, action checks, format conversion, subtask segmentation, and caption generation. Captions include action descriptions at three levels of detail, along with scene and lighting information.}
\label{fig:data-curation-pipeline}
\end{figure}

In Figure~\ref{fig:data-curation-pipeline}, we summarize the curation process used to turn the four experience sources into training-ready video--language examples. We perform quality checks, standardize the video records, segment long videos into subtasks where applicable, and generate multi-level captions before mixing the resulting data for pretraining.

\noindent\textbf{General videos.} We use OpenVid~\cite{DBLP:conf/iclr/NanXZFYCL0T25} and Panda70M~\cite{DBLP:conf/cvpr/ChenSMDCJF0RYT24}, totaling 10,909.6 hours. These open-source video generation datasets contain videos and text descriptions covering diverse scenes, objects, and visual dynamics. We retain their videos and original descriptions to provide language--video experience beyond manipulation demonstrations.

\noindent\textbf{Egocentric videos.} We use Ego4D~\cite{grauman2022ego4d}, EgoVerse~\cite{punamiya2026egoverse}, and EgoSuite~\cite{lightwheel2026egosuite}, totaling 371,212 trajectories and 10,986.3 hours. Ego4D primarily records daily activities across real-life settings. EgoVerse covers more diverse scenes, adding egocentric visual experience across environments. EgoSuite is currently the largest open-source egocentric manipulation video dataset. We retain the videos and language descriptions from these sources.

\noindent\textbf{Data preparation.} We package all sources in LeRobot 2.1 format. We retain only samples whose videos, language descriptions, and metadata can be linked. We exclude samples with unassociated video, language, or metadata, videos with encoding or decoding failures, and known low-quality samples. Video records retain their corresponding language descriptions and metadata. The inventory counts the recorded episodes and videos at their source durations.

\noindent\textbf{Visual content filtering.} For general video datasets such as OpenVid and Panda70M, we use original annotations and VLM-based content review to remove animations, static scenes, television programs, and other unsuitable content. For egocentric videos, we use a VLM to assess motion blur, brightness, occlusion, and whether a clip contains meaningful actions, reducing the impact of low-quality imagery and uninformative behavior on training. The UMI and robot demonstrations described below use similar filtering rules.

\noindent\textbf{Video model adaptation on general videos.} Our WAM modifies
Wan's video modeling design to support robot control, which may disrupt
pretrained visual and dynamics priors. To ease this adaptation, we first train
the video branch on general videos, whose distribution is close to that of
Wan's pretraining data. During this stage, we use the same input frame count,
spatial resolution, and causal attention scheme as subsequent WAM training.
This stage aims to preserve these priors during adaptation, before introducing
embodied data.

\subsubsection{Language Annotation}
\label{sec:method:language-annotation}

Language annotations describe manipulated objects and executed actions and may additionally include interaction relationships, manipulation processes, and environmental information. For WAM pretraining, what semantic information helps translate video experience into control capabilities? Can these annotations help preserve language--video alignment and improve downstream generalization? We organize language supervision along three dimensions: the level of detail in manipulation descriptions, environmental information, and temporal granularity.

\noindent\textbf{Level of detail.} Task descriptions cover executed actions, interaction objects, and their attributes during manipulation. We define three levels of descriptive detail, Brief, Medium, and Detailed, with requirements and examples in Table~\ref{tab:language-detail}. These levels define controlled language-annotation variants for evaluating how description detail affects the transfer of pretrained video capabilities to robot control.

\begin{table}[htbp]
\centering\small
\caption{\textbf{Language annotation requirements and example descriptions at three levels of detail.} All examples describe the same manipulation segment.}
\label{tab:language-detail}
\setlength{\tabcolsep}{4pt}
\begin{tabular}{p{.12\linewidth}p{.36\linewidth}p{.43\linewidth}}
\toprule
\textbf{Detail level} & \textbf{Annotation requirements} & \textbf{Example language annotation} \\
\midrule
Brief & Summarize the main action and core object with a verb--object phrase. & Pick up the snack. \\
Medium & Summarize the manipulation in one sentence, adding object attributes, spatial relations, or visible outcomes. & Pick up the snack in the light-blue package. \\
Detailed & Describe the manipulation chronologically, including discernible hand actions, object interactions, and necessary adjustments. & The right hand first touches and grasps the snack in the light-blue package, then lifts it from the table, while the left hand does not participate. \\
\bottomrule
\end{tabular}
\end{table}

The three descriptions convey the same manipulation. The Brief level specifies the main action and object, the Medium level adds the object's appearance, and the Detailed level further specifies hand involvement and action order.

\noindent\textbf{Scene information.} We use Qwen 3.8 Max \cite{bai2025qwen3} and Seed 2.1 Pro \cite{seed2026seed2} for subtask segmentation and caption annotation. Scene annotations supplement task descriptions with the working environment, spatial layout, and initial state. Lighting annotations describe brightness, exposure, illumination uniformity, and shadows. For example, the scene annotation for one clip is: ``The manipulation takes place on a table covered with a black-and-white striped tablecloth, with various packaged snacks scattered across the surface.'' Its lighting annotation is: ``Indoor lighting is bright and uniform, with no obvious shadows or overexposed regions. All snacks and the manipulation area are clearly visible.''

\noindent\textbf{Temporal granularity.} Episode descriptions cover an entire trajectory, while Subtask descriptions correspond to local manipulation segments with start and end times. A VLM segments continuous video into subtasks and outputs each segment's timestamps and description. Descriptive detail and temporal granularity are independent annotation dimensions: the same subtask can have descriptions at different detail levels, and an entire trajectory can also have a brief or detailed description. During training, the Episode setting uses the description of the full trajectory, while the Subtask setting uses the annotation of the corresponding time interval as the language condition.

\Needspace{5\baselineskip}
\subsection{Language--Video--Action Alignment}

UMI and robot demonstrations follow the video curation and filtering procedures in Section~\ref{sec:method:video-curation} and use the language annotations described in Section~\ref{sec:method:language-annotation}. Building on these language--video correspondences, we align video segments with state and action records to construct language--video--action supervision. We describe the collection and curation of these demonstrations below.

\noindent\textbf{UMI demonstrations.} We use 411,754 in-house UMI trajectories totaling 15,284.9 hours. We collect human manipulation demonstrations using RoboPocket \cite{fang2026robopocket}. The system integrates a smartphone with a handheld adaptive gripper into a portable collection device. Operators interact with objects by moving, opening, and closing the gripper. Its structure and interaction mechanism resemble robot end effectors, reducing interface differences between human demonstrations and robot execution. Our collection tasks include mouse placement, block sorting, seasoning pouring, towel folding, bimanual snack bagging, and other everyday manipulation tasks. They involve pick-and-place, long-horizon manipulation, large wrist rotations, deformable object manipulation, and bimanual coordination. Collection environments include indoor and outdoor scenes with different lighting and background textures.

During collection, head and bilateral wrist views record the overall manipulation scene and local interaction details, respectively. Smartphone visual--inertial tracking provides action trajectories, while external encoders record gripper opening and closing. A shared coordinate system and time reference establish spatial and temporal correspondences among the videos of both hands, motion trajectories, and gripper states. The collection procedure includes equipment checks and synchronization verification before recording, tracking stability and kinematic feasibility checks during recording, and video and AR trajectory playback afterward. These checks identify anomalies such as localization drift and trajectory jumps and verify task completion.

\noindent\textbf{Robot demonstrations.} We curate data from AgiBot World Alpha/Beta/2026~\cite{bu2025agibot}, RoboCOIN~\cite{wu2025robocoin}, RoboMIND~\cite{wu2024robomind}, InternData-A1~\cite{tian2026interndata}, Galaxea Open-World~\cite{jiang2026galaxy}, and ABC-130K~\cite{allshire2026scalable}. These sources span different robot embodiments, camera configurations, and manipulation environments. We retain only demonstration subsets collected with bimanual robots that include head and bilateral wrist views and meet our quality requirements, maintaining consistent data interfaces and observation semantics. We exclude subsets with missing views, inconsistent view semantics, insufficient image resolution, overly simple tasks, single-arm robot embodiments, or other issues. After applying the configured episode exclusion lists, we retain 768,356 trajectories totaling 9,836.0 hours.

\noindent\textbf{State--action representation.} UMI and robot demonstrations use a shared 16-dimensional bimanual end-effector representation. Each hand has a three-dimensional position, a four-dimensional rotation quaternion in wxyz order, and a one-dimensional gripper value. State and action records are associated with their corresponding video segments. Each source is normalized using its own q1 and q99 statistics.

\noindent\textbf{Trajectory quality.} For our in-house UMI demonstrations, we construct action-supervision masks from the state and action validity annotations provided by the collection pipeline. Videos participate in video learning, while valid action labels additionally provide action supervision and invalid labels are masked. For the collected robot demonstrations, we first check action dimensions, missing values, NaNs, non-finite values, and quaternion validity. For subsets with the required joint and end-effector state fields, we further exclude trajectories with abnormal joint angles, excessive changes in joint angles or end-effector poses between adjacent frames, or no movement throughout the trajectory. Trajectories that pass these quality checks are used for action training.


\begingroup
\definecolor{capEpisodeBg}{HTML}{EEEAF7}
\definecolor{capEpisodeFg}{HTML}{67538E}
\definecolor{capSubtaskBg}{HTML}{E7F1F8}
\definecolor{capSubtaskFg}{HTML}{267DA6}
\definecolor{capBriefBg}{HTML}{FBEFE8}
\definecolor{capBriefFg}{HTML}{9A5735}
\definecolor{capMediumBg}{HTML}{F8F1DD}
\definecolor{capMediumFg}{HTML}{876B2B}
\definecolor{capDetailedBg}{HTML}{EAF3E8}
\definecolor{capDetailedFg}{HTML}{476A42}
\definecolor{capEnvironmentBg}{HTML}{E5F2EF}
\definecolor{capEnvironmentFg}{HTML}{39776A}
\newcommand{\capcell}[2]{\begingroup\setlength{\fboxsep}{1pt}\colorbox{#1Bg}{\textcolor{#1Fg}{#2}}\endgroup}
\newcommand{\fullrowcolor}[1]{\noalign{%
\begingroup\color{#1}\hrule height\dimexpr\ht\csname @arstrutbox\endcsname+\dp\csname @arstrutbox\endcsname\relax\endgroup
\vskip-\dimexpr\ht\csname @arstrutbox\endcsname+\dp\csname @arstrutbox\endcsname\relax}}
\setlength{\emergencystretch}{2em}
\titleformat{\paragraph}[block]{\normalfont\normalsize\bfseries}{}{0pt}{#1}
\titlespacing*{\paragraph}{0pt}{1.5ex plus 0.5ex minus 0.2ex}{0.8ex}

\section{Experiments}
\label{sec:experiments}
We aim to use multisource experience to preserve the language--video
capabilities acquired during video pretraining and extend them to
generalizable robot control. Our experiments examine how language--video alignment,
language--video--action alignment, joint training, and pretraining scale
affect downstream generalization. We organize our study around the following four questions.
\begin{itemize}
  \item How to establish effective cross-modal alignment?
  \item How to integrate multisource experience?
  \item Does scaling pretraining improve control generalization?
  \item How well does our model perform in unseen environments?
\end{itemize}

\subsection{Experimental Setup}
\label{sec:exp:setup}

\titleformat{\paragraph}[runin]{\normalfont\normalsize\bfseries}{}{0pt}{#1}
\titlespacing*{\paragraph}{0pt}{1.5ex plus 0.5ex minus 0.2ex}{0.5em}
\paragraph{Pretraining.}
We pretrain the model using the architecture and multisource corpus
introduced above. The video branch is initialized from
Wan2.2-TI2V-5B~\cite{wan2025wan}, and the action branch is randomly
initialized. All model parameters are updated during training.
Pretraining and post-training both use the joint video and action
objective defined above. We use AdamW with an initial learning rate of
$4\times10^{-5}$ and a default global batch size of 256. Training steps
and specific settings for each study are given in the corresponding
experiment descriptions and table captions.

\paragraph{Simulation post-training.}
We use the 50 manipulation tasks in RoboTwin 2.0~\cite{chen2025robotwin}.
For the main simulation comparisons, we include VLA baselines
(StarVLA~\cite{community2026starvla}, $\pi_{0.5}$~\cite{intelligence2025pi_},
Xiaomi Robotics-0~\cite{cai2026xiaomi}, and X-VLA~\cite{zheng2026x}) and
video-based WAM baselines (Fast-WAM~\cite{yuan2026fast},
LingBot-VA~\cite{li2026causal}, OpenWAM~\cite{yu2026openwam},
X-WAM~\cite{guo2026unified}, and 4D-WAM~\cite{yang20264d}).
Observations include head and bilateral wrist-camera views and the
end-effector state. The model predicts bimanual end-effector actions.
Each task provides 50 Clean demonstrations, yielding 2,500 demonstrations
in total. Starting from the corresponding pretrained checkpoint, we
jointly post-train a single model on all 50 tasks for 20k steps with
a global batch size of 128.

\paragraph{Simulation evaluation.}
We evaluate control generalization on RoboTwin under two settings.
Models are post-trained exclusively on Clean demonstrations.
\textbf{Easy} evaluates the resulting models under the same clean conditions
used for post-training (Clean-to-Clean, \textbf{C2C}). \textbf{Hard} evaluates the same models
under randomized conditions, including variations in background textures,
clutter, lighting, table height, and camera placement (Clean-to-Random, \textbf{C2R}).
For each task and setting, we run 25 evaluation episodes for ablation
studies and 100 episodes for main results. We report full-task success
rate (SR).

\paragraph{Real-world post-training.}\label{sec:exp:real-world}
We use a bimanual Flexiv Rizon 4 platform for cup stacking, towel folding,
and gift-in-box.
For real-world evaluation, we select StarVLA, $\pi_{0.5}$, Fast-WAM,
LingBot-VA, and OpenWAM as representative VLA and WAM baselines.
Observations comprise two fisheye wrist-camera views
and one fixed head-camera view. All compared models use the same
task-specific post-training data.
Starting from upstream pretrained weights, we adapt a separate model
to each task for 50k steps with a global batch size of 128.

\paragraph{Real-world evaluation.}
We evaluate under \textbf{Easy (C2C)} and \textbf{Hard (C2R)} settings. Easy varies
the positions of familiar objects, and Hard changes objects,
backgrounds, or both. Each task and setting uses ten layouts fixed
before evaluation and shared across all models. We run three trials
per layout, yielding 30 trials per task and setting. All models use
synchronous inference and execute actions at 30\,Hz.
Success rate (SR) measures the proportion of trials that complete the entire task.
Partial success rate (PSR) measures the average fraction of predefined subtasks completed per trial.
We compute both metrics per task and average them equally across the three tasks.

\titleformat{\paragraph}[block]{\normalfont\normalsize\bfseries}{}{0pt}{#1}
\titlespacing*{\paragraph}{0pt}{1.5ex plus 0.5ex minus 0.2ex}{0.8ex}

\subsection{How to establish effective cross-modal alignment?}
\label{sec:exp:alignment}
We examine language--video and language--video--action alignment through
the design of language annotations, paired action supervision, and
complementary video experience. Downstream control evaluation measures
the effects of these supervision choices.

\subsubsection{Language--Video Alignment}
\label{sec:exp:text-video}

\noindent\textbf{\studyhl{swamHLLanguage}{Detailed and accurate language descriptions improve downstream control generalization.}}\expfindingtoc{Rich and aligned language supervision improves generalization}\label{sec:exp:caption-content}
Pretrained video models learn rich information about scenes, objects,
and motion from large-scale language--video data. During WAM pretraining, we provide informative language supervision
that accurately matches video content. We aim to help the model retain
pretrained priors and reduce forgetting of existing language--video
capabilities when adapting to robot manipulation data. Across all settings, we pretrain on a mixture of egocentric videos and UMI
demonstrations for 20k steps, examining description detail, scene information,
and temporal granularity as three aspects of language supervision.
For description detail, we compare three levels, Brief, Medium, and Detailed.
We also compare settings with and without scene descriptions covering
backgrounds, spatial layouts, and lighting. For temporal granularity,
we compare Episode descriptions covering complete trajectories with
Subtask descriptions of local manipulation segments.

We compare these language supervision choices in Table~\ref{tab:exp:captions}.
At each temporal granularity, more detailed descriptions yield higher
Hard success rates. At the Subtask level, the Hard success rate increases
from 47.20\% to 49.60\% and 53.52\% with Brief, Medium, and Detailed
descriptions, respectively. Adding environment information further raises
it to 55.04\%, achieving the highest Easy and Hard success rates in this
group. For temporal granularity, replacing Episode descriptions with
Subtask descriptions under the Detailed setting increases the Hard
success rate from 51.60\% to 53.52\%.
These results indicate that richer manipulation descriptions, additional
environment information, and subtask-level language supervision improve
downstream control generalization.
\label{sec:exp:temporal}We further randomly shuffle the correspondence
between Detailed Subtask descriptions and video segments within each
trajectory, reducing the Hard success rate from 53.52\% to 52.00\%.
This result further highlights the importance of accurately matched
language supervision for language--video alignment.

\begin{table}[!htbp]
\centering
\begin{minipage}[t]{0.50\linewidth}
\vspace{0pt}\centering\small
\setlength{\tabcolsep}{4pt}
\caption{\textbf{\studyhl{swamHLLanguage}{Caption design and temporal alignment.}} We vary description detail, scene information, temporal granularity, and caption--video correspondence. \studyhl{black!15}{Gray rows} indicate the configurations used in our final model.}
\label{tab:exp:captions}\label{tab:exp:temporal}
\begin{tabular}{llcc}
    \toprule
    \textbf{Granularity} & \textbf{Setting} & \textbf{Easy (C2C)} & \textbf{Hard (C2R)} \\
    \midrule
    \multicolumn{4}{l}{\textit{Caption design}} \\
\hspace{0.8em}Episode & Brief & 81.36 & 44.96 \\
\hspace{0.8em}Episode & Detailed & 82.00 & 51.60 \\
\hspace{0.8em}Subtask & Brief & 83.60 & 47.20 \\
\hspace{0.8em}Subtask & Medium & 82.08 & 49.60 \\
\hspace{0.8em}Subtask & Detailed & 83.68 & 53.52 \\
\fullrowcolor{black!10}
\hspace{0.8em}Subtask & Detailed + Env. & \textbf{83.92} & \textbf{55.04} \\
    \midrule
    \multicolumn{4}{l}{\textit{Temporal alignment}} \\
\hspace{0.8em}Subtask & Aligned & \textbf{83.68} & \textbf{53.52} \\
\hspace{0.8em}Subtask & Shuffled & 83.04 & 52.00 \\
    \bottomrule
\end{tabular}
\end{minipage}\hfill
\begin{minipage}[t]{0.46\linewidth}
\vspace{0pt}\centering\small
\captionsetup{skip=4pt}
\caption{\textbf{\studyhl{swamHLLanguage}{Robot caption variants.}} We compare original task instructions, rephrased instructions, and detailed trajectory-specific descriptions.}
\label{tab:exp:caption-variants}
{\fontsize{8.5}{9.5}\selectfont
\begin{tabular}{lcc}
\toprule
\textbf{Caption} & \textbf{Easy (C2C)} & \textbf{Hard (C2R)} \\
\midrule
Original & 84.64 & 50.80 \\
Rephrased & 86.08 & 54.76 \\
Detailed + Env. & \textbf{86.48} & \textbf{56.16} \\
\bottomrule
\end{tabular}
}
\par\vspace{6pt}
\caption{\textbf{\studyhl{swamHLAction}{UMI action-label coverage.}} All videos are retained as action-label coverage varies.}
\label{tab:exp:action-supervision}
{\fontsize{8.5}{9.5}\selectfont
\begin{tabular}{ccc}
\toprule
\textbf{Action} & \textbf{Easy (C2C)} & \textbf{Hard (C2R)} \\
\midrule
0\% & 80.96 & 46.72 \\
25\% & 81.44 & 47.76 \\
50\% & 83.28 & 49.92 \\
75\% & \textbf{84.08} & 50.16 \\
100\% & 83.28 & \textbf{51.28} \\
\bottomrule
\end{tabular}
}
\end{minipage}
\end{table}

We also find that many trajectories in the open-source robot data we examine share
the same simple task instruction. Some datasets use task names or short instructions
as video captions, without detailed descriptions of the manipulation process and
environment. For this study, we randomly select nine subsets of the robot data,
containing 145,430 trajectories but only 592 task-level instruction entries. These
shared instructions do not adequately capture differences in manipulation processes
and scene conditions across trajectories, limiting fine-grained correspondence
between language supervision and trajectory-level visual content. We construct two
variants to enrich language supervision. Rephrased provides 100 semantically
equivalent paraphrases per task instruction, randomly sampling one during training to
increase linguistic diversity. Detailed + Env. provides a detailed description of the
specific manipulation process and environment for each trajectory, further enriching
the content and diversity of language supervision. Relative to the 50.80\% Hard
success rate of Original, paraphrasing alone yields a gain of 3.96 percentage points,
while Detailed + Env. achieves the highest success rate in this group at 56.16\%, as
shown in Table~\ref{tab:exp:caption-variants}. These results further indicate that
enriching linguistic expression and descriptive content helps improve
language--video alignment and generalization.

\subsubsection{Language--Video--Action Alignment}
\label{sec:exp:text-video-action}

\noindent\textbf{\studyhl{swamHLAction}{Human manipulation experience provides effective language--video--action supervision.}}\expfindingtoc{Human manipulation experience provides action supervision}\label{sec:exp:action-supervision}
Language–video supervision associates language descriptions with manipulation
processes in videos, while paired action labels further provide supervision for
language--video--action alignment. Direct post-training from Wan achieves a Hard SR of 34.56\%.
With robot-data pretraining, it reaches 56.16\% in Table~\ref{tab:exp:caption-variants},
demonstrating the value of robot manipulation data. We further examine whether human manipulation experience from UMI
can provide effective cross-embodiment language--video--action supervision. To
examine the contribution of UMI action supervision, we retain all UMI videos and vary
only the proportion of trajectories providing action labels during pretraining.
Videos without action labels continue to participate in video learning. At each
coverage level, the subset of trajectories providing action supervision remains
fixed. All settings use the same pretraining budget, followed by 20k steps of
downstream post-training. From Table~\ref{tab:exp:action-supervision}, we observe
that as action-label coverage increases from 0\% to 100\%, the Hard success rate
increases across all five coverage levels, rising from 46.72\% to 51.28\%, a gain of
4.56 percentage points. These results indicate that paired actions from human
manipulation experience provide effective supervision for robot control, although UMI
data differ visually from robot execution. Given that UMI demonstrations can be
collected with handheld devices across diverse scenes, this finding supports human
manipulation demonstrations as a source for scaling pretraining experience for robot
control.


\subsection{How to integrate multisource experience?}
\label{sec:exp:distributions}
After examining the roles of language--video and language--video--action supervision, we
study how to integrate multisource experience for pretraining. We first
examine the benefits of increasing training on General videos, egocentric videos,
and UMI demonstrations. Given differences across sources in data quality, scene
and task distributions, and action supervision, we then study how the Ego--UMI
sampling ratio affects downstream control generalization under a fixed
training budget. Finally, we compare model architectures and conditioning
designs under mixed-source training, examining how the pretrained model's
existing language--video capabilities can support action learning.

\begin{table}[!htbp]
\centering\small
\caption{\textbf{\studyhl{swamHLSources}{Different data sources strengthen language--video and language--video--action alignment.}} We separately increase pretraining exposure to general videos (\textbf{General}), egocentric videos (\textbf{Ego}), and UMI data (\textbf{UMI}).}
\label{tab:exp:ego-scaling}
\begin{tabular}{lrcc}
\toprule
\textbf{Data} & \textbf{Steps} & \textbf{Easy (C2C)} & \textbf{Hard (C2R)} \\
\midrule
\multicolumn{4}{l}{\textit{General videos}} \\
\hspace{0.8em}UMI/Ego 20k & 20k & \textbf{82.64} & 53.84 \\
\hspace{0.8em}+ General 25k & 45k & \textbf{82.64} & \textbf{56.48} \\
\midrule
\multicolumn{4}{l}{\textit{Egocentric videos}} \\
\hspace{0.8em}UMI only & 10k & \textbf{83.04} & 49.68 \\
\hspace{0.8em}+ Ego 10k & 20k & 82.64 & 53.84 \\
\hspace{0.8em}+ Ego 20k & 30k & 82.32 & \textbf{55.60} \\
\midrule
\multicolumn{4}{l}{\textit{UMI data}} \\
\hspace{0.8em}Ego only & 10k & 79.68 & 42.08 \\
\hspace{0.8em}+ UMI 10k & 20k & 82.64 & \textbf{53.84} \\
\hspace{0.8em}+ UMI 20k & 30k & \textbf{83.60} & 51.20 \\
\bottomrule
\end{tabular}
\end{table}
\FloatBarrier

\noindent\textbf{\studyhl{swamHLSources}{Transitional pretraining on General videos provides a stronger starting point.}}\expfindingtoc{Transitional pretraining on General videos provides a stronger starting point}\label{sec:exp:video-stage}
We first examine the role of pretraining on General videos. Their content
distribution is close to that of the original video pretraining data, while WAM
adjusts input frame count, resolution, and attention computation to support robot
control. We therefore first use General videos to help the model adapt to these
computational changes under a content distribution close to its pretraining
distribution. This transition aims to better preserve existing video priors and
provide a foundation for subsequent embodied learning. To this end, we add 25k steps
of transitional pretraining using only General videos before 20k steps of mixed
training on egocentric videos and UMI demonstrations, increasing the Hard success
rate from 53.84\% to 56.48\%, as reported in Table~\ref{tab:exp:ego-scaling}.

\noindent\textbf{\studyhl{swamHLSources}{Increasing pretraining exposure to individual data sources improves WAM performance.}}\expfindingtoc{Pretraining exposure to individual data sources improves WAM performance}\label{sec:exp:source-exposure}\label{sec:exp:ego-scaling}
Egocentric videos broaden the range of egocentric manipulation scenarios, and UMI
demonstrations provide manipulation experience paired with actions. We next increase
training exposure to egocentric videos or UMI demonstrations while keeping exposure
to the other source fixed. We mix egocentric videos and UMI demonstrations during
training. Each source's training exposure is calculated from the total training steps
and its sampling proportion. As shown in Table~\ref{tab:exp:ego-scaling}, increasing
training on egocentric videos or UMI demonstrations generally improves downstream
control performance.

\begin{wraptable}{R}{0.38\textwidth}
\centering\small
\captionsetup{skip=6pt}
\caption{\textbf{\studyhl{swamHLSampling}{Ego and UMI provide complementary supervision.}} We compare five Ego-to-UMI sampling ratios under a fixed 20k-step pretraining budget.}
\label{tab:exp:sources}
{\small\setlength{\tabcolsep}{8pt}
\begin{tabular}{ccc}
\toprule
\textbf{Ratio} & \textbf{Easy (C2C)} & \textbf{Hard (C2R)} \\
\midrule
1:0 & 77.04 & 42.56 \\
2:1 & 82.72 & 52.64 \\
\fullrowcolor{black!10}
1:1 & 82.64 & \textbf{53.84} \\
1:2 & 83.20 & 53.28 \\
0:1 & \textbf{83.28} & 51.28 \\
\bottomrule
\end{tabular}
}
\end{wraptable}

\noindent\textbf{\studyhl{swamHLSampling}{Equal Ego--UMI sampling yields the best performance in unseen environments.}}\expfindingtoc{Equal Ego--UMI sampling gives the best generalization}\label{sec:exp:sources}
The preceding experiments increase both source exposure and the training
budget. We further examine how to allocate a fixed budget across the two
sources. Egocentric videos provide diverse visual experience across tasks and scenes, while UMI data provide both
video and action supervision. The sampling ratio determines how these forms of
supervision are allocated during training. We fix the pretraining budget at
20k steps and compare five Ego-to-UMI sampling ratios of 1:0, 2:1, 1:1, 1:2,
and 0:1. In Table~\ref{tab:exp:sources}, we observe that all three mixed settings
achieve higher Hard success rates than training on either source alone. Equal
sampling achieves the highest Hard success rate of 53.84\%, improving over
UMI-only training at 51.28\% by 2.56 percentage points. These results indicate
that an appropriate Ego-to-UMI sampling ratio helps balance language--video
and language--video--action supervision and improves control generalization in
randomized scenes.

\begin{wraptable}{R}{0.38\textwidth}
\vspace{-\intextsep}
\centering\small
\captionsetup{skip=4pt}
\caption{\textbf{\studyhl{swamHLArchitecture}{Model architecture affects control generalization.}} We compare video--action architectures (top) and extra image features (bottom).}
\label{tab:exp:architecture}\label{tab:exp:visual}
{\small\setlength{\tabcolsep}{3.5pt}
\begin{tabular}{lcc}
\toprule
 & \textbf{Easy (C2C)} & \textbf{Hard (C2R)} \\
\midrule
\multicolumn{3}{l}{\textit{Architecture}} \\
\hspace{0.8em}Single-stream & \textbf{86.48} & 52.56 \\
\hspace{0.8em}Dual-stream & 84.96 & 52.48 \\
\fullrowcolor{black!10}
\hspace{0.8em}Ours & 84.80 & \textbf{57.36} \\
\midrule
\multicolumn{3}{l}{\textit{Extra features}} \\
\fullrowcolor{black!10}
\hspace{0.8em}Native & \textbf{83.44} & \textbf{49.52} \\
\hspace{0.8em}+ CLIP~\cite{radford2021learning} & 82.40 & 42.48 \\
\hspace{0.8em}+ DINOv2~\cite{DBLP:journals/tmlr/OquabDMVSKFHMEA24} & 82.00 & 39.76 \\
\hspace{0.8em}+ SigLIP2~\cite{tschannen2025siglip} & 82.40 & 42.40 \\
\hspace{0.8em}+ Qwen3-VL & 81.60 & 42.96 \\
\bottomrule
\end{tabular}
}
\end{wraptable}

\noindent\textbf{\studyhl{swamHLArchitecture}{Appropriate architectural choices help preserve pretrained video priors.}}\expfindingtoc{Architecture helps preserve pretrained video priors}\label{sec:exp:architecture}\label{sec:exp:streams} Under the
same 30k-step multisource pretraining setup, we compare Single-stream, in which video
and actions share a residual stream, with Dual-stream, which maintains
separate residual streams. We also examine the language conditioning pathway
of the action branch. We separate video and action feature spaces to accommodate both the reuse of pretrained video representations and the learning of new action representations. The video branch has already
learned correspondences between language and visual dynamics during
pretraining. The newly introduced text conditioning pathway in the action
branch, however, lacks sufficient pretraining. We therefore aim to reuse the
semantic information available in the video branch for action prediction,
reducing potential interference from the new text conditioning pathway. To
this end, we retain language conditioning in the video branch and let the
action branch access language information only through video features.
As reported in Table~\ref{tab:exp:architecture}, Single-stream and standard
Dual-stream achieve similar Hard success rates of 52.56\% and 52.48\%,
respectively. The dual-stream model using this language conditioning pathway
achieves the highest Hard success rate in this group at 57.36\%, a gain of
4.88 percentage points over standard Dual-stream. These results support
conditioning the action branch on language through video features.

\label{sec:exp:encoders}\label{sec:exp:first-frame}For visual conditioning, we
examine whether image-encoder features can provide benefits. We pretrain for 20k steps with equal sampling of egocentric videos and UMI data. Using Native as the baseline,
we separately add image features extracted by CLIP, DINOv2, SigLIP2, and
Qwen3-VL 2B~\cite{bai2025qwen3} through cross-attention to provide additional
visual conditioning. Native achieves the highest Easy and Hard success rates
in this group, at 83.44\% and 49.52\%, respectively. Adding features from the
four image encoders lowers the Hard success rate to 39.76\%--42.96\%,
as summarized in Table~\ref{tab:exp:visual}. These results support retaining the pretrained
video model's native visual inputs. The fusion pathway for the additional image features has not undergone
the same large-scale language--video pretraining as the video branch. This may
lead the model to rely more on appearance cues in the current training
distribution, weakening generalization to unseen
scenes.

\subsection{Does scaling pretraining improve control generalization?}
\label{sec:exp:scalability}

Building on the preceding supervision design and multisource training
recipe, we evaluate different pretrained checkpoints in simulation and
on real robots. Within each evaluation domain, we use the same downstream
post-training settings to examine how additional pretraining affects
control generalization.

\noindent\textbf{\studyhl{swamHLScaling}{Increasing pretraining investment in multisource experience provides a stronger WAM.}}\expfindingtoc{Scaling pretraining improves WAM performance}\label{sec:exp:data-scaling}\label{sec:exp:real-pretraining}
We study how model capabilities scale with pretraining, measured in training video hours.
Here, 50k pretraining steps correspond to approximately 45k video hours.
We use a mixture of egocentric videos and UMI demonstrations with equal sampling weights.
For this study, we combine the configurations marked by gray rows in Tables~\ref{tab:exp:captions}, \ref{tab:exp:sources}, and~\ref{tab:exp:architecture}, such as using Detailed Subtask descriptions with scene and lighting information as language conditions.
The 0-hour setting skips multisource pretraining and proceeds directly to downstream
post-training from the video-model initialization.

\begin{figure}[!htbp]
\centering
\includegraphics[width=\linewidth]{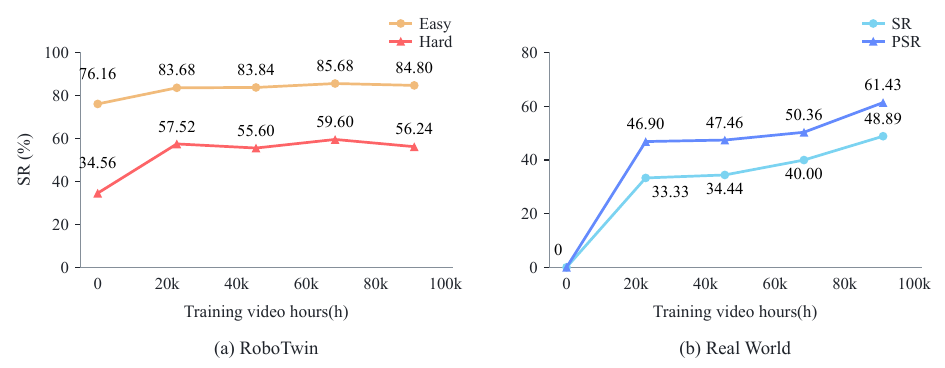}
\caption{\textbf{\studyhl{swamHLScaling}{Scaling pretraining improves control generalization.}} We vary training video hours on a fixed corpus of multisource experience, using the same downstream post-training protocol within each domain. (a) RoboTwin SR under the Easy (C2C) and Hard (C2R) settings. (b) Real-world SR and PSR under the Hard setting, averaged across three tasks. The 0-hour setting uses the video-model initialization without additional pretraining and achieves no success in real-world evaluation.}
\label{fig:exp:real-pretraining}
\end{figure}

As shown in Figure~\hyperref[fig:exp:real-pretraining]{\ref*{fig:exp:real-pretraining}(a)}, multisource pretraining improves both Easy and Hard success rates over the 0-hour baseline. The highest Hard success rate reaches 59.60\%, compared with 34.56\% without pretraining.
In Figure~\hyperref[fig:exp:real-pretraining]{\ref*{fig:exp:real-pretraining}(b)}, we show that real-world Hard SR and PSR,
averaged equally across three tasks, increase consistently as training video hours grow from 22.5k to 90k, rising from 33.33\% and 46.90\% to 48.89\% and
61.43\%, respectively. The 0-hour setting fails to complete
any task. These results show continued gains within the evaluated range,
supporting longer pretraining on both sources for control under real-world scene
changes.


\begin{table}[!htbp]
\centering\small
\caption{\textbf{\studyhl{swamHLScaling}{Scaling pretraining on human experience and robot data improves generalization across unseen environmental conditions.}} We first pretrain on egocentric videos and UMI demonstrations (\textbf{Stage 1}), then continue training on robot data (\textbf{Stage 2}). The middle four columns evaluate single-factor environmental changes.}
\label{tab:exp:condition-comparison}
\begin{tabular}{rrcrrrrc}
\toprule
\textbf{Stage 1} & \textbf{Stage 2} & \textbf{Easy (C2C)} & \textbf{Background} & \textbf{Lighting} & \textbf{Clutter} & \textbf{Height} & \textbf{Hard (C2R)} \\
\midrule
0k & 0k & 76.16 & 59.12 & 76.24 & 51.28 & 57.68 & 34.56 \\
20k & 0k & 82.64 & 72.16 & 84.08 & 68.56 & 70.32 & 53.84 \\
75k & 0k & 85.68 & 77.92 & 85.60 & 78.48 & 75.44 & 59.60 \\
75k & 25k & \textbf{88.08} & \textbf{82.16} & \textbf{88.08} & \textbf{80.72} & \textbf{76.96} & \textbf{61.52} \\
\bottomrule
\end{tabular}
\end{table}

\label{sec:exp:stages}After 75k steps of pretraining on egocentric videos and UMI demonstrations, we continue
pretraining on robot data for another 25k steps and apply the same
RoboTwin post-training settings. In Table~\ref{tab:exp:condition-comparison}, we observe that success rates
under individual changes in background, lighting, clutter, and table
height improve as pretraining investment increases, with the Hard
success rate ultimately reaching 61.52\%.

\endgroup

\begingroup
\setlength{\emergencystretch}{2em}
\subsection{Does our model perform well in unseen environments?}
\label{sec:exp:main-comparison}

Building on the preceding findings and training recipe, we compare our models pretrained on our full multisource dataset with existing policies in RoboTwin simulation and real-world manipulation.

\noindent\textbf{Simulation results.}\expfindingtoc{Simulation results} We compare VLA and WAM methods on
the 50 manipulation tasks in RoboTwin 2.0. Each model is post-trained on Clean demonstrations and evaluated under Easy (C2C) and Hard (C2R) settings. As reported in Table~\ref{tab:exp:baseline-summary}, our model achieves a
Hard success rate of 61.56\%, outperforming all compared methods and
exceeding the strongest baseline, OpenWAM, by 12.86 percentage points. Its Easy success rate is 88.10\%,
slightly below OpenWAM's 89.30\%, while its average success rate across
both settings reaches 74.83\%, the highest among the compared methods.
Our model and OpenWAM perform similarly under Clean conditions, while our
model achieves a higher overall success rate under the Hard
setting, indicating better generalization across scene conditions.

\begin{table}[!htbp]
\centering\small
\captionsetup{font=small}
\caption{\textbf{Our model outperforms VLA and WAM baselines in simulation and real-world evaluation.} We compare Easy (C2C) and Hard (C2R) settings, reporting SR in simulation and SR/PSR on real-world tasks.}
\label{tab:exp:baseline-summary}
\setlength{\tabcolsep}{3pt}
\begin{tabular*}{\linewidth}{@{\extracolsep{\fill}\hspace{\tabcolsep}}lccccccc}
\toprule
\multirow{2}{*}{\textbf{Model}} & \multicolumn{3}{c}{\textbf{RoboTwin}} & \multicolumn{4}{c}{\textbf{Real-world}} \\
\cmidrule(lr){2-4}\cmidrule(l){5-8}
 & \textbf{\textcolor{black}{Easy}} & \textbf{\textcolor{black}{Hard}} & \textbf{Avg.} & \textbf{\textcolor{black}{Easy} SR} & \textbf{\textcolor{black}{Easy} PSR} & \textbf{\textcolor{black}{Hard} SR} & \textbf{\textcolor{black}{Hard} PSR} \\
\midrule
\multicolumn{8}{l}{\textit{VLA}} \\
\hspace{0.8em}StarVLA~\cite{community2026starvla} & 60.03 & 14.06 & 37.05 & 31.11 & 36.87 & 4.44 & 21.55 \\
\hspace{0.8em}$\pi_{0.5}$~\cite{intelligence2025pi_} & 70.66 & 46.00 & 58.33 & 50.00 & \textbf{71.87} & 20.00 & 37.78 \\
\hspace{0.8em}Xiaomi Robotics-0~\cite{cai2026xiaomi} & 62.86 & 18.22 & 40.54 & -- & -- & -- & -- \\
\hspace{0.8em}X-VLA~\cite{zheng2026x} & 67.96 & 20.95 & 44.46 & -- & -- & -- & -- \\
\midrule
\multicolumn{8}{l}{\textit{WAM}} \\
\hspace{0.8em}Fast-WAM~\cite{yuan2026fast} & 77.76 & 1.92 & 39.84 & 35.56 & 49.33 & 6.67 & 14.52 \\
\hspace{0.8em}LingBot-VA~\cite{li2026causal} & 77.04 & 36.16 & 56.60 & 41.11 & 53.61 & 10.00 & 24.01 \\
\hspace{0.8em}OpenWAM~\cite{yu2026openwam} & \textbf{89.30} & 48.70 & 69.00 & 37.78 & 46.35 & 23.33 & 31.67 \\
\hspace{0.8em}X-WAM~\cite{guo2026unified} & 69.98 & 25.84 & 47.91 & -- & -- & -- & -- \\
\hspace{0.8em}4D-WAM~\cite{yang20264d} & 81.50 & 41.80 & 61.65 & -- & -- & -- & -- \\
\midrule
\hspace{0.8em}Ours & 88.10 & \textbf{61.56} & \textbf{74.83} & \textbf{56.67} & 68.57 & \textbf{40.00} & \textbf{50.36} \\
\bottomrule
\end{tabular*}
\end{table}

\noindent\textbf{Real-world results.}\expfindingtoc{Real-world results} We evaluate cup stacking, towel
folding, and gift-in-box on a bimanual Flexiv platform. All three tasks start from the same pretrained checkpoint and undergo 50k steps of task-specific post-training. Each task-specific model is evaluated under
both Easy (C2C) and Hard (C2R) settings. All compared methods use the same
task-specific post-training data. Under the Hard setting, our model achieves the highest SR and PSR among the evaluated methods, as shown in Table~\ref{tab:exp:baseline-summary} and Figure~\ref{fig:exp:baseline-comparison}.
Averaged equally across the three tasks, SR reaches 40.00\%,
exceeding OpenWAM's 23.33\% by 16.67
percentage points. PSR reaches 50.36\%, exceeding $\pi_{0.5}$'s
37.78\% by 12.58 percentage points. These results show improvements in
both full-task success and subtask completion under unseen object and
background conditions, supporting the generalization benefits of the
training recipe in real-world robot manipulation.

\begin{figure}[!htbp]
\centering
\includegraphics[width=\linewidth]{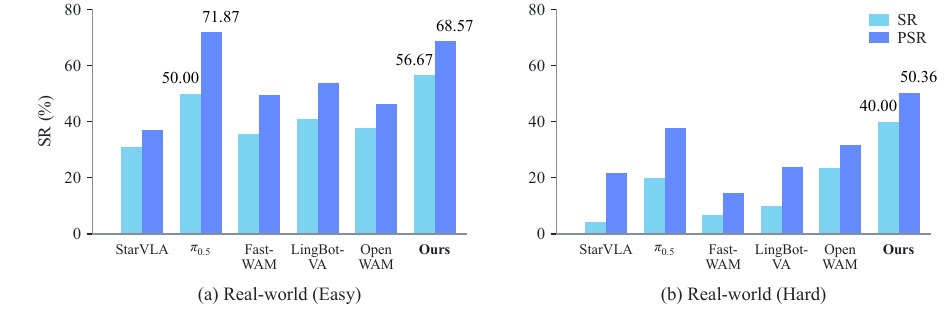}
\caption{\textbf{Our method achieves the best performance under the Hard setting in real-world evaluation.} Results are averaged equally across three real-world tasks.}
\label{fig:exp:baseline-comparison}
\end{figure}

\paragraph*{Real-world visualizations.}\expfindingtoc{Real-world visualizations}
Figure~\ref{fig:exp:real-execution} shows how our model and the baselines perform the three tasks, with light red dashed boxes highlighting the baselines' main execution errors.
In cup stacking, Fast-WAM misplaces the first cup of the second layer, causing it to fall onto the cups below, and cannot complete the stack afterward.
For towel folding, our model lifts and spreads both sides before completing both folds, whereas OpenWAM lifts only one side and fails to fold the towel properly.
In gift-in-box, $\pi_{0.5}$ grasps the object but gradually loosens its gripper, causing the object to fall, and remains unsuccessful after repeated attempts.
Our model retains its grasp and places the object into the box.
To provide a fuller view of our model's execution, Figure~\ref{fig:exp:real-views} presents head and external camera recordings.
The gift-in-box model generalizes to fruit placement with a different scene, manipulated objects, and target container.
Trained with same-color cups on a gray tablecloth, the cup-stacking model also completes rainbow stack with multicolored cups on a patterned tablecloth.
These examples illustrate how our model transfers learned manipulation skills to unfamiliar objects and scenes.

\begin{figure}[!htbp]
\centering
\includegraphics[width=0.8\linewidth]{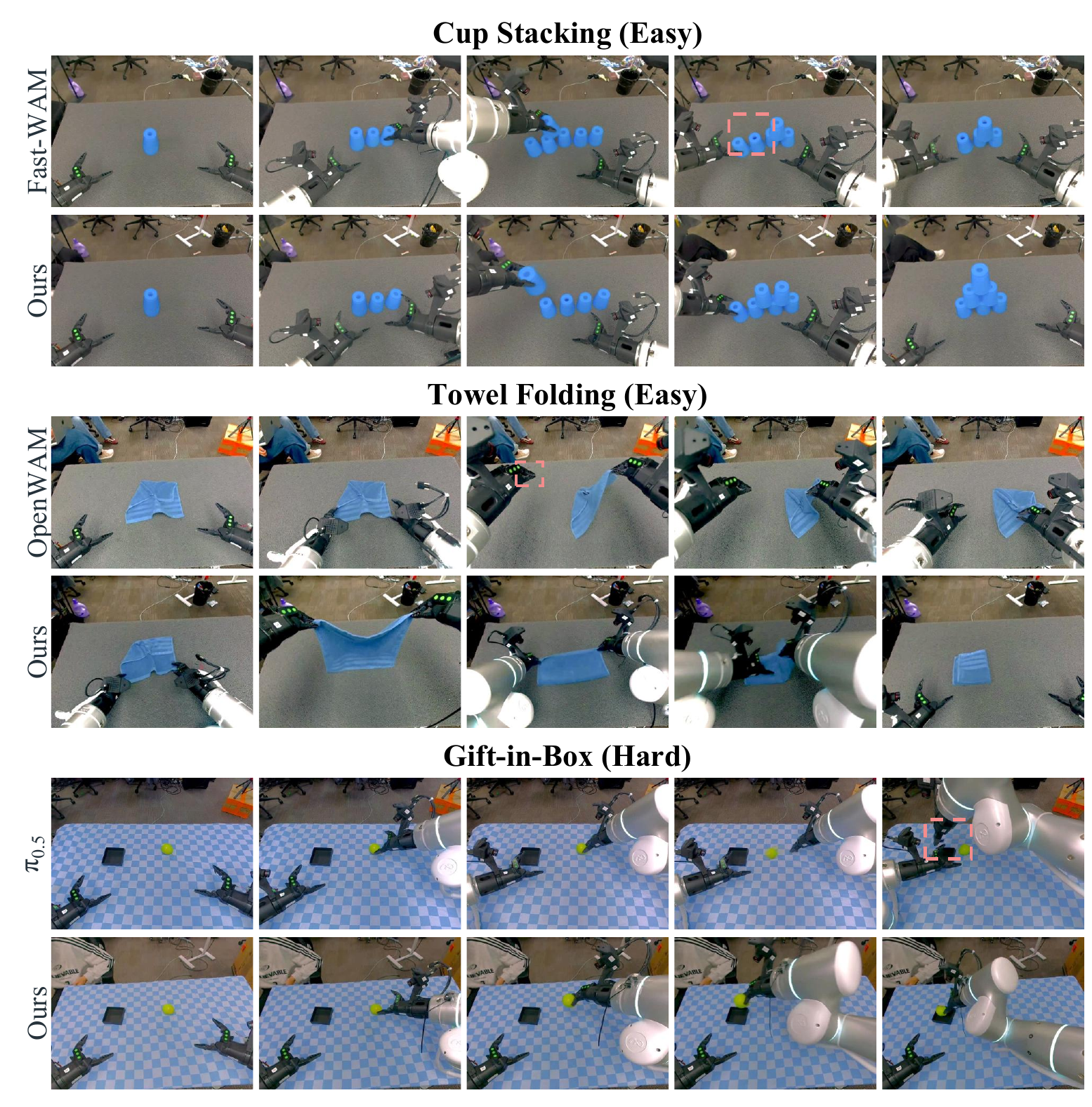}
\caption{\textbf{Real-world execution comparisons.} Our model completes cup stacking and towel folding under the Easy setting and gift-in-box under the Hard setting, while the corresponding baseline executions fail.
\textcolor{red!55}{Light red dashed boxes} highlight the baselines' main execution errors.}
\label{fig:exp:real-execution}
\end{figure}

\begin{figure}[!htbp]
\centering
\includegraphics[width=0.8\linewidth]{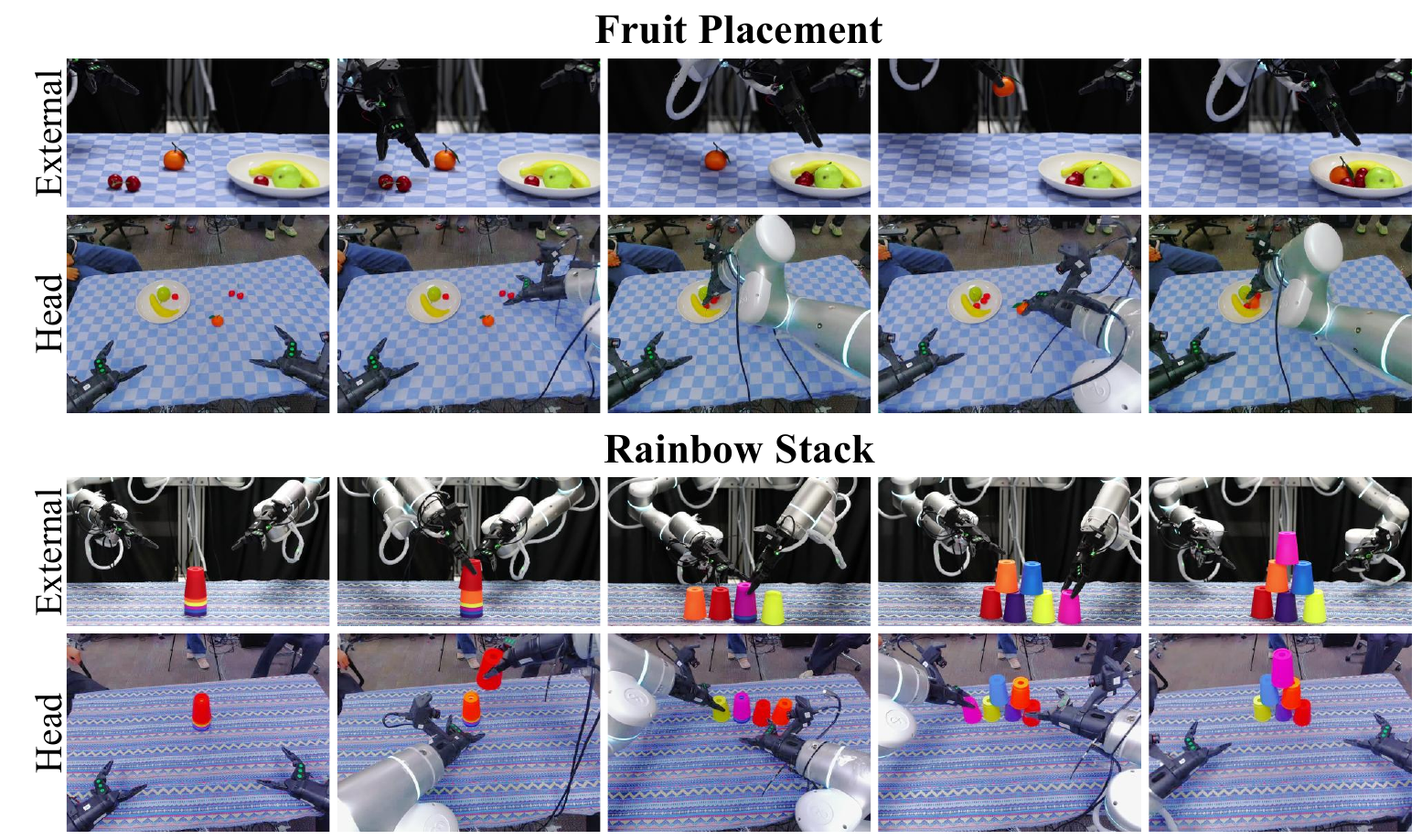}
\caption{\textbf{Generalization examples from head and external camera views.}
Fruit placement and rainbow stack demonstrate generalization to new objects, containers, and scene appearances.
}
\label{fig:exp:real-views}
\end{figure}
\endgroup

\section{Discussion and Conclusion}
\enlargethispage{\baselineskip}

In this work, we studied how to extend pretrained language--video
capabilities to generalizable robot control with multisource experience.
Using approximately 47,000 hours
of experience, we systematically investigated language--video alignment,
language--video--action alignment, and their joint learning. 
We found detailed and accurate language descriptions preserve pretrained language--video capabilities. 
Meanwhile, action supervision from UMI demonstrations further supports robot control, demonstrating the value of human manipulation experience for robot learning. 
Our findings prove the importance of experience organization and supervision usage. 
The resulting training recipe demonstrates generalization to unseen conditions in simulation and real-world evaluations, with further gains from continued pretraining. 
We release the data, code, and model weights to support future research.

Upon the pretrained video model, the proposed method offers limited support for complex reasoning and explicit multi-stage planning, despite outperforming VLA models in our control generalization evaluations. 
The limited historical context may also hinder the execution of long-range tasks. 
In the future, we plan to explore memory mechanisms and hierarchical planning to support tasks that require reasoning over past observations and planning subsequent actions.

\section*{Acknowledgments}

We thank Zijun Xu for his contributions to the early exploration of this work.
We also thank Yutong Dong for her efforts in real-world deployment.

\clearpage
\bibliographystyle{template/swam-unsrt}
\bibliography{main}

\clearpage
\appendix
\phantomsection
\addcontentsline{toc}{section}{\protect\numberline{6}Appendix}
\begingroup
\let\appendixaddcontentsline\addcontentsline
\renewcommand{\addcontentsline}[3]{%
  \ifstrequal{#1}{toc}{}{\appendixaddcontentsline{#1}{#2}{#3}}%
}
\section{Data Annotation Details}
\label{app:data-annotation-details}
General videos retain their original captions. For egocentric videos, UMI
demonstrations, and robotic demonstrations, we construct manipulation annotations
using Qwen 3.8 Max primarily for subtask segmentation and Seed 2.1 Pro for
caption annotation. The procedure consists of the following steps.

\noindent\textbf{Subtask Segmentation.} We use Qwen 3.8 Max to divide
egocentric videos and UMI recordings into temporally ordered, non-overlapping
subtasks according to action intent, recording each subtask's start and end
time. Robotic demonstrations are short, so each complete episode is treated as
one subtask without further segmentation.\par

\noindent\textbf{Caption Annotation.} We use Seed 2.1 Pro to annotate each
egocentric or UMI subtask and each robotic episode. The model identifies
whether the segment contains manipulation and whether the visual evidence is
sufficiently clear, and describes the scene, lighting, and actions at multiple
levels of detail. Segments that contain no manipulation or whose visual evidence is insufficient are discarded entirely.\par

\noindent\textbf{Action Quality Filtering.} For UMI and robotic data, we further screen the associated state and action trajectories. Samples that fail one or more of the following checks are removed from action training:
\begin{itemize}
  \item Missing or non-finite values, including NaNs, invalid dimensions, or invalid quaternions.
  \item Excessive absolute joint values, with any $\lvert q\rvert > 20\,\mathrm{rad}$.
  \item Adjacent-frame joint jumps, with any $\lvert q_{t+1}-q_t\rvert > 1\,\mathrm{rad}$.
  \item End-effector position jumps, with the Euclidean distance between adjacent 3D positions exceeding $0.2\,\mathrm{m}$.
  \item End-effector orientation jumps, with a rotation-angle difference exceeding $90^\circ$ for either arm between adjacent frames.
  \item Persistent abnormal jerk in the joint or end-effector trajectories.
\end{itemize}
\par

Condensed versions of the segmentation and caption annotation prompts are
provided below.

\begin{tcolorbox}[enhanced,breakable,colback=white,colframe=swamBlue,
  colbacktitle=swamBlue,coltitle=white,title=\textbf{Split Prompt (Condensed)},
  fonttitle=\bfseries,boxrule=0.6pt,arc=1mm,left=3mm,right=3mm,
  top=2mm,bottom=2mm]
\small
\textbf{Role.} You are a professional robot manipulation video analysis expert.

\textbf{Goal.} Split a video into
continuous, non-overlapping atomic subtasks. The input video is
sampled at 2 fps, with 0.5 seconds between consecutive frames. The overall task is: \texttt{task\_name}.

\textbf{Workflow.}
\begin{enumerate}
  \item Watch the complete video and identify every meaningful atomic action in
  chronological order.
  \item Split by action intent.
  \item Ensure that subtasks are sequential, with no overlap or gap.
  \item Name objects accurately.
\end{enumerate}

\textbf{Output.} Output standardized JSON containing the description, start time, and end time of every subtask.

\begin{verbatim}
[
  {
    "description": "subtask description",
    "start_time": 0.500,
    "end_time": 2.500
  }
]
\end{verbatim}
\end{tcolorbox}

\begin{tcolorbox}[enhanced,breakable,colback=white,colframe=swamBlue,
  colbacktitle=swamBlue,coltitle=white,title=\textbf{Caption Prompt (Condensed)},
  fonttitle=\bfseries,boxrule=0.6pt,arc=1mm,left=3mm,right=3mm,
  top=2mm,bottom=2mm]
\small
\textbf{Role.} You are a professional first-person manipulation video analysis
expert and dense action annotation specialist. Describe scenes and continuous manipulation actions accurately, objectively, and
concisely.

\textbf{Task.} Watch the input video and use only visually observable evidence.
The reference task, \texttt{task\_annotation: \{task\_name\}}, is only a
weak hint; all descriptions must be supported by the visible video content.

\textbf{Segment-type classification.} The input is a temporal segment cut from
a complete video and may contain no actual manipulation, for example waiting,
posture adjustment, hovering, or observation. First determine:
\begin{itemize}
  \item \texttt{segment\_type = manipulation} when the segment contains an
  actual operation on a task object.
  \item \texttt{segment\_type = no\_action} when the segment contains no
  operation on a task object.
\end{itemize}

\textbf{Visual-clear classification.} The field \texttt{visual\_clear} only
describes the quality of the visual evidence:
\begin{itemize}
  \item Set \texttt{visual\_clear = true} when the main objects and actions
  are sufficiently visible to judge, even with ordinary motion blur or local
  occlusion.
  \item Set \texttt{visual\_clear = false} when severe blur, darkness,
  occlusion, corruption, or missing views make the main objects or actions
  unreliable to judge.
\end{itemize}

\textbf{Scene.} Describe the initial operating environment in one
 sentence. Mention only the visible workspace, relevant background
  objects, approximate positions, and initial state.

\textbf{Lighting.} Describe brightness, exposure, uniformity, shadows,
overexposure, underexposure, blur, or occlusion in one sentence.

\textbf{action\_detail.} Describe the manipulation as one chronological paragraph, including manipulated objects, discernible hand actions, object trajectories, interactions, state changes, and necessary adjustments.

\textbf{action\_medium.} Describe the manipulation in one sentence, adding object attributes, spatial relations, or visible outcomes.

\textbf{action\_short.} Summarize the main action and core object with a 2--8 word verb-object phrase.

\textbf{Output.} Return only one valid JSON object with the following fields.
Use exactly \texttt{manipulation} or \texttt{no\_action} for
\texttt{segment\_type}, and a JSON boolean for \texttt{visual\_clear}.
Write descriptions in English. For \texttt{no\_action}, set the three action
fields to \texttt{none}; otherwise describe only observed actions.
\begin{verbatim}
{
  "segment_type": "manipulation",
  "visual_clear": true,
  "lighting": "",
  "scene": "",
  "action_detail": "",
  "action_medium": "",
  "action_short": ""
}
\end{verbatim}

\end{tcolorbox}

\section{Training Details}
\label{app:training-details}

\subsection{Common Settings}
\label{app:common-training-settings}

\paragraph{Optimization.}
\label{app:common-optimization}
Pretraining and post-training share the optimization settings in Table~\ref{tab:app:common-optimization}.
Batch sizes, training steps, and learning-rate schedules vary across training stages, as detailed in Section~\ref{app:training-configurations}.

\begin{table}[!htbp]
\centering\small
\caption{\textbf{Shared optimization settings.} Values shared across pretraining and post-training.}
\label{tab:app:common-optimization}
\setlength{\tabcolsep}{12pt}
\begin{tabular}{cc}
\toprule
\textbf{Configuration} & \textbf{Value} \\
\midrule
Optimizer & AdamW \\
Learning rate & $4\times10^{-5}$ \\
Adam $(\beta_1,\beta_2)$ & $(0.95,\,0.999)$ \\
Adam $\epsilon$ & $10^{-8}$ \\
Weight decay & $10^{-5}$ \\
Gradient clip & 1.0 \\
Warmup ratio & 5\% \\
Random seed & 42 \\
\bottomrule
\end{tabular}
\end{table}

\paragraph{Model inputs and trainable components.}
\label{app:common-model-architecture}
We use three camera views. Each view is resized to $128\times160$ and arranged into a 2$\times$2 grid, yielding a final $256\times320$ composite frame. The head-camera view is duplicated across the top row, while the left- and right-wrist views occupy the bottom row. During training, each view is randomly cropped to 95\% of its original extent before resizing, followed by color jitter with brightness, contrast, saturation, and hue parameters of 0.3, 0.4, 0.5, and 0.08, respectively.
We represent bimanual end-effector states and actions as 16-dimensional vectors and predict actions in chunks of 48 steps.

\subsection{Training Configurations}
\label{app:training-configurations}

\paragraph{Initialization.}
In pretraining, we initialize the video DiT from Wan2.2-TI2V-5B~\cite{wan2025wan} and load its pretrained UMT5-XXL text encoder and Wan2.2 video VAE.
The action branch, including the action and state encoders, transformer blocks, and action decoder, is randomly initialized.
For each experiment, post-training starts from the corresponding pretrained weights.

\paragraph{Pretraining.}
\label{app:pretraining-recipe}
Training uses BF16 mixed precision.
The default global batch size for ablation studies is 256.
We increase it to 512 for pretraining in the scaling study and subsequent robot-data pretraining described in Section~\ref{sec:exp:scalability}, as well as for pretraining the final models evaluated in Section~\ref{sec:exp:main-comparison}.
For the scaling study, we sample egocentric videos and UMI demonstrations with equal probability and use detailed subtask descriptions with scene and lighting information.
We evaluate checkpoints at 25k, 50k, 75k, and 100k steps, as shown in Figure~\ref{fig:exp:real-pretraining}.
Table~\ref{tab:app:training-configurations} compares pretraining in the scaling study with simulation and real-world post-training.

\paragraph{Simulation post-training.}
\label{app:robotwin-posttraining-recipe}
We jointly train a single model on all 50 RoboTwin 2.0 tasks using 50 Clean demonstrations per task, following the same post-training protocol across simulation experiments.
Each sample contains left-wrist, right-wrist, and head-camera views and the current bimanual end-effector state. The same three-view preprocessing and augmentation settings described above are used during simulation post-training.
We use absolute end-effector states and actions and normalize both with their q99 statistics.

\paragraph{Real-world post-training.}
\label{app:real-posttraining-recipe}
For each real-world task, we train an independent model using task-level instructions.
We apply this procedure to cup stacking, towel folding, and gift-in-box.

\begin{table}[!htbp]
\centering\small
\caption{\textbf{Comparison of training configurations.}}
\label{tab:app:training-configurations}
\setlength{\tabcolsep}{10pt}
\begin{tabular}{lccc}
\toprule
\textbf{Configuration} & \textbf{Pretraining} & \multicolumn{2}{c}{\textbf{Post-training}} \\
\cmidrule(lr){2-2}\cmidrule(l){3-4}
 & Scaling study & Simulation & Real world \\
\midrule
Global batch size & 512 & 128 & 128 \\
Training steps & 100,000 & 20,000 & 50,000 \\
Learning-rate schedule & Constant & Cosine & Cosine \\
\bottomrule
\end{tabular}
\end{table}

\section{Real-World Evaluation Protocol}
\label{app:real-protocol}
\subsection{Post-Training Data Preparation}
\label{app:real-data-preparation}

\paragraph{Post-training data.}
We adapt models separately to cup stacking, towel folding, and gift-in-box using task-specific data for post-training following general-purpose pretraining on egocentric videos and UMI demonstrations.
Table~\ref{tab:app:real-data} summarizes the post-training data for each task.

\begin{table}[ht]
\centering\small
\caption{\textbf{Task-specific post-training data for real-world evaluation.}}
\label{tab:app:real-data}
\setlength{\tabcolsep}{12pt}
\begin{tabular}{lrr}
\toprule
Task & Episodes & Hours \\
\midrule
Cup stacking & 2,748 & 25.788 \\
Towel folding & 2,852 & 15.112 \\
Gift-in-box & 4,399 & 21.715 \\
\bottomrule
\end{tabular}
\end{table}

\begin{figure}[ht]
\centering
\includegraphics[width=0.8\linewidth]{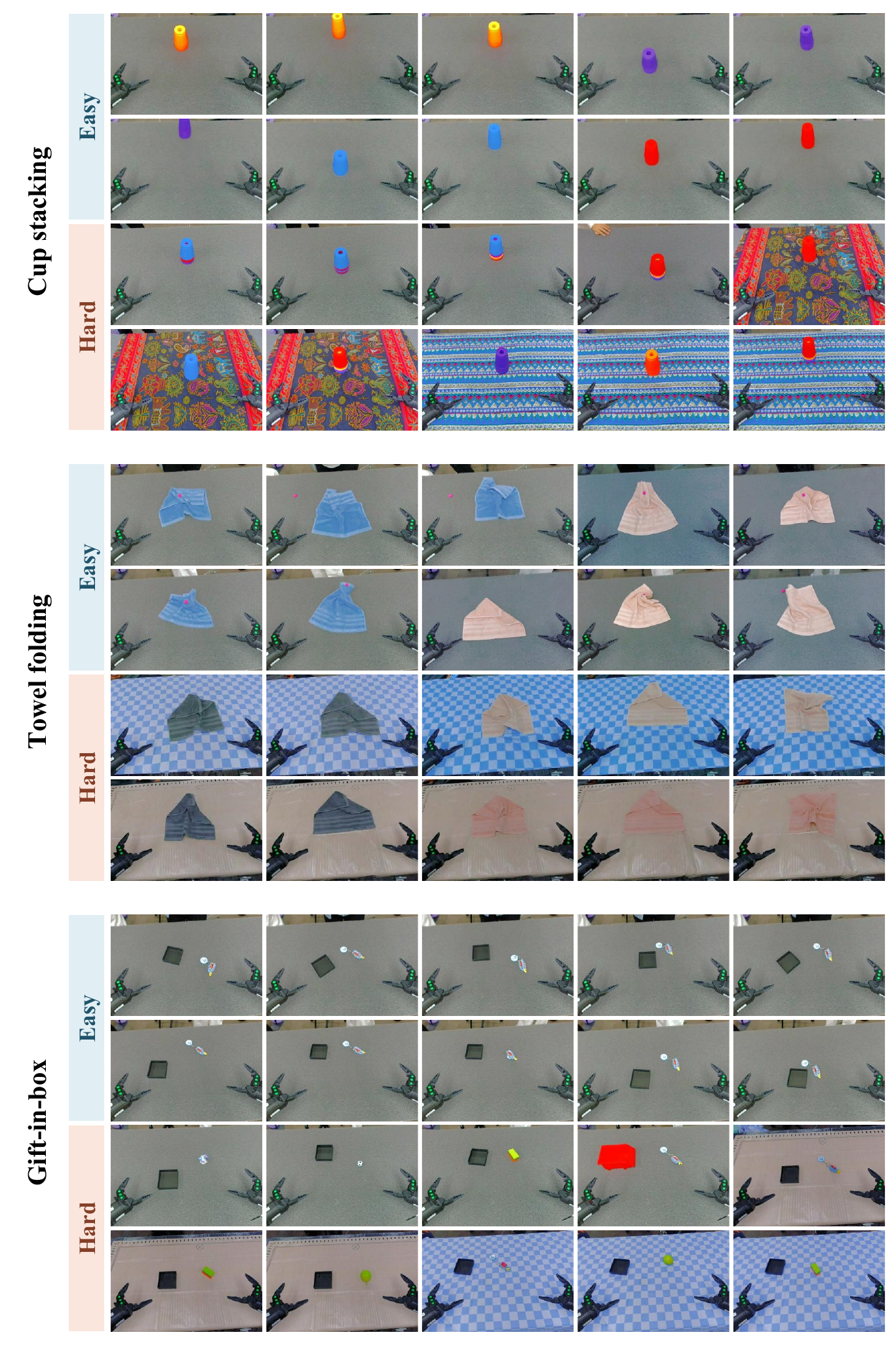}%
\caption{\textbf{Real-world evaluation layouts under Easy and Hard settings.}
All models use the same ten layouts for each task and setting, with three trials per layout.
}
\label{fig:app:real-layouts}
\end{figure}

\paragraph{Observations and actions.}
We use a bimanual Flexiv Rizon 4 platform.
Observations comprise two fisheye wrist-camera views and one fixed head-camera view.
Actions encode the end-effector position, orientation, and continuous gripper width for each arm.
For OpenWAM, we follow its official default configuration and use a 6D rotation representation.
All other models represent orientation with a wxyz quaternion, yielding a 16-dimensional action vector across both arms.

\begin{figure}[!htbp]
\centering
\includegraphics[width=0.8\linewidth]{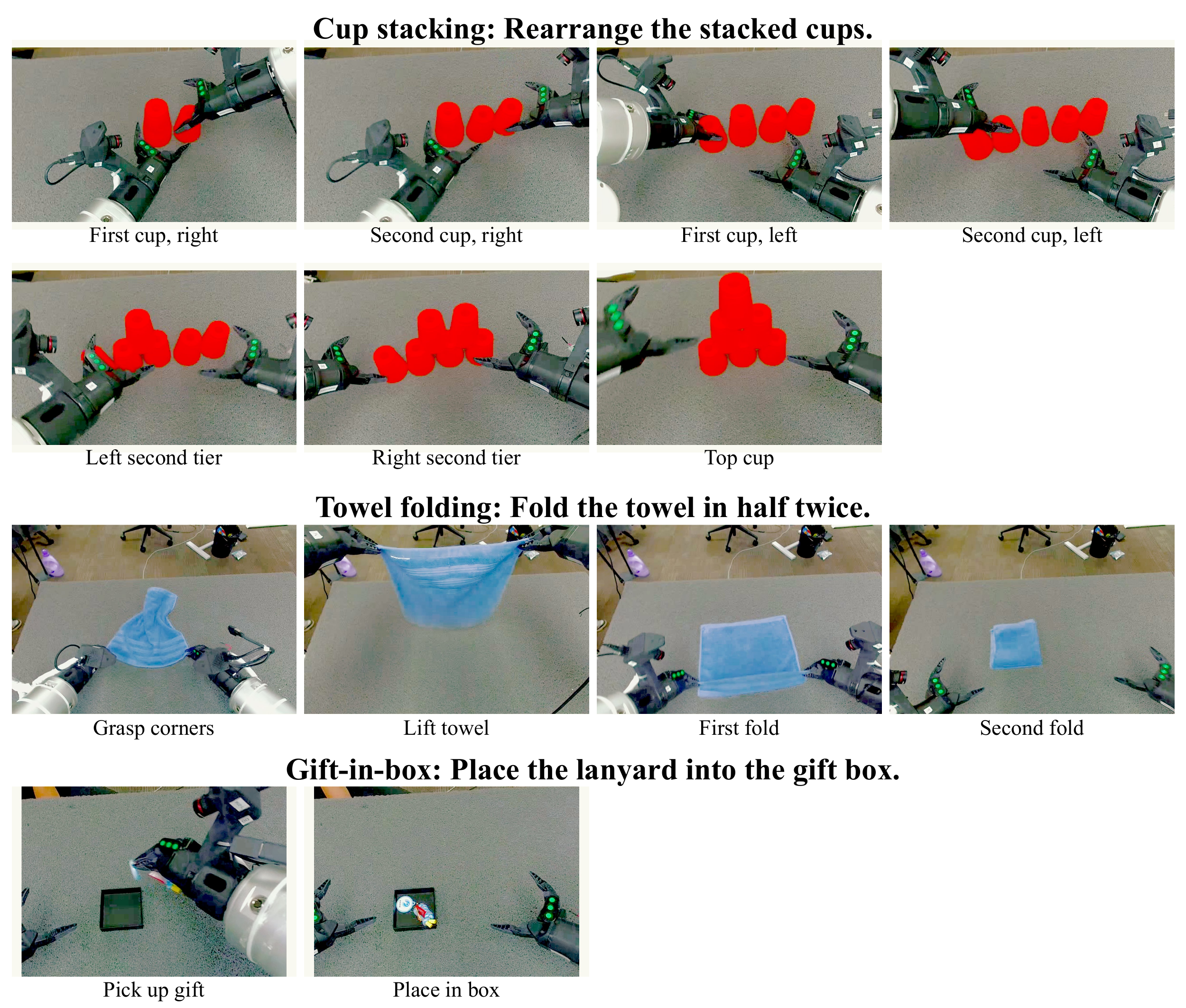}
\caption{\textbf{Subtask illustrations for the three real-world tasks.}}
\label{fig:app:real-subtasks}
\end{figure}

\FloatBarrier

\subsection{Evaluation Protocols and Metrics}
\label{app:real-evaluation}

\paragraph{Evaluation protocols.}
We evaluate under \textbf{Easy (C2C)} and \textbf{Hard (C2R)} settings.
Easy settings retain the scene conditions and physical objects used in post-training, including their shapes and colors, while varying their initial configurations across layouts.
Hard settings vary across scene and object attributes, including background, tabletop material, object instance, shape, color, size, and texture.
The evaluation layouts are illustrated in Figure~\ref{fig:app:real-layouts}.
All models use synchronous inference and execute actions at 30\,Hz.

\paragraph{Layouts and trials.}
A layout specifies the initial object configurations, target positions, and scene appearance.
For each task and setting, ten layouts are fixed before evaluation and shared across all models.
We run three trials per layout to account for variability in object placement and model sampling, yielding 30 trials per task and setting.
To reproduce a layout, we align the fixed live head-camera view with a reference photograph using an image overlay.

\paragraph{Scoring and termination.}
Each task is divided into predefined, equally weighted subtasks, scored by reviewing execution videos.
The protocol permits up to five additional autonomous replanning attempts after an initial failed attempt, retaining the current environment state and execution history without human assistance.
All models follow the same retry and termination rules.
A trial is marked as a failure if the cup tower collapses or an object leaves the workspace.

\paragraph{Metrics.}
We report success rate (SR) and partial success rate (PSR).
For a task with $N=30$ trials and $K$ subtasks, let $s_{ik}\in\{0,1\}$ indicate whether subtask $k$ succeeds in trial $i$.
Complete-task success requires all subtasks to succeed, so its indicator is
$z_i=\prod_{k=1}^{K}s_{ik}$.
The two metrics are
\begin{equation}
\mathrm{SR}=\frac{1}{N}\sum_{i=1}^{N}z_i,
\qquad
\mathrm{PSR}=\frac{1}{NK}\sum_{i=1}^{N}\sum_{k=1}^{K}s_{ik}.
\label{eq:app:real-metrics}
\end{equation}
SR measures complete-task success. PSR measures progress at the subtask level.
The three-task score is the equally weighted mean of the task-level rates, computed separately for Easy and Hard and expressed as a percentage.

\paragraph{Task details.}
\label{app:real-tasks}

Figure~\ref{fig:app:real-subtasks} shows the three real-world tasks and their corresponding ordered subtask decompositions.
Cup stacking requires arranging cups into a stable pyramid, towel folding requires folding a towel in half twice, and gift-in-box requires placing a gift inside an open gift box.
We divide these tasks into seven, four, and two subtasks, respectively, corresponding to observable milestones in task execution.
These decompositions define the task-specific subtask counts $K$ used in the PSR calculation above.

\FloatBarrier

\section{Detailed Results}
\label{app:detailed-results}
\subsection{RoboTwin Task Success Rates}
\label{app:robotwin-results}

\begin{table}[!htbp]
\centering\small
\caption{\textbf{Per-task success rates on RoboTwin 2.0 under Easy and Hard settings.} Each task is evaluated over 100 episodes per setting. All values are percentages.}
\label{tab:app:robotwin-tasks}
\begingroup
\setlength{\tabcolsep}{4pt}
\renewcommand{\arraystretch}{1.15}
\begin{tabular*}{\linewidth}{@{\extracolsep{\fill}\hspace{\tabcolsep}}lrr@{\hspace{22pt}}lrr}
\toprule
\textbf{Task} & \textbf{Easy} & \textbf{Hard} & \textbf{Task} & \textbf{Easy} & \textbf{Hard} \\
\midrule
Adjust Bottle & 97 & 81 & Place Can Basket & 84 & 36 \\
Beat Block Hammer & 97 & 55 & Place Cans Plasticbox & 99 & 74 \\
Blocks Ranking RGB & 98 & 80 & Place Container Plate & 96 & 87 \\
Blocks Ranking Size & 87 & 30 & Place Dual Shoes & 56 & 33 \\
Click Alarmclock & 100 & 75 & Place Empty Cup & 100 & 77 \\
\addlinespace[3pt]
Click Bell & 100 & 85 & Place Fan & 87 & 48 \\
Dump Bin Bigbin & 91 & 66 & Place Mouse Pad & 73 & 29 \\
Grab Roller & 100 & 80 & Place Object Basket & 84 & 56 \\
Handover Block & 75 & 31 & Place Object Scale & 94 & 67 \\
Handover Mic & 78 & 40 & Place Object Stand & 98 & 83 \\
\addlinespace[3pt]
Hanging Mug & 3 & 1 & Place Phone Stand & 87 & 28 \\
Lift Pot & 100 & 56 & Place Shoe & 98 & 82 \\
Move Can Pot & 90 & 72 & Press Stapler & 81 & 78 \\
Move Pillbottle Pad & 95 & 58 & Put Bottles Dustbin & 90 & 82 \\
Move Playingcard Away & 98 & 56 & Put Object Cabinet & 78 & 57 \\
\addlinespace[3pt]
Move Stapler Pad & 58 & 22 & Rotate QRcode & 90 & 50 \\
Open Laptop & 93 & 81 & Scan Object & 95 & 57 \\
Open Microwave & 86 & 4 & Shake Bottle & 100 & 98 \\
Pick Diverse Bottles & 94 & 74 & Shake Bottle Horizontally & 99 & 98 \\
Pick Dual Bottles & 99 & 85 & Stack Blocks Three & 93 & 82 \\
\addlinespace[3pt]
Place A2B Left & 92 & 52 & Stack Blocks Two & 100 & 99 \\
Place A2B Right & 95 & 59 & Stack Bowls Three & 84 & 80 \\
Place Bread Basket & 87 & 67 & Stack Bowls Two & 99 & 90 \\
Place Bread Skillet & 92 & 57 & Stamp Seal & 94 & 29 \\
Place Burger Fries & 97 & 60 & Turn Switch & 44 & 51 \\
\bottomrule
\end{tabular*}
\endgroup
\end{table}

\FloatBarrier
\subsection{Real-World Task Success Rates}
\label{app:real-results}

\subsubsection{Baseline Comparison}
\label{app:real-baselines}
Figure~\ref{fig:real-baselines-detailed} reports SR and PSR for each real-world task and their average under the Easy and Hard settings.

\begin{figure}[!htbp]
    \centering
    \captionsetup{skip=5pt}
    \includegraphics[width=\textwidth,trim=0 12bp 0 0,clip]{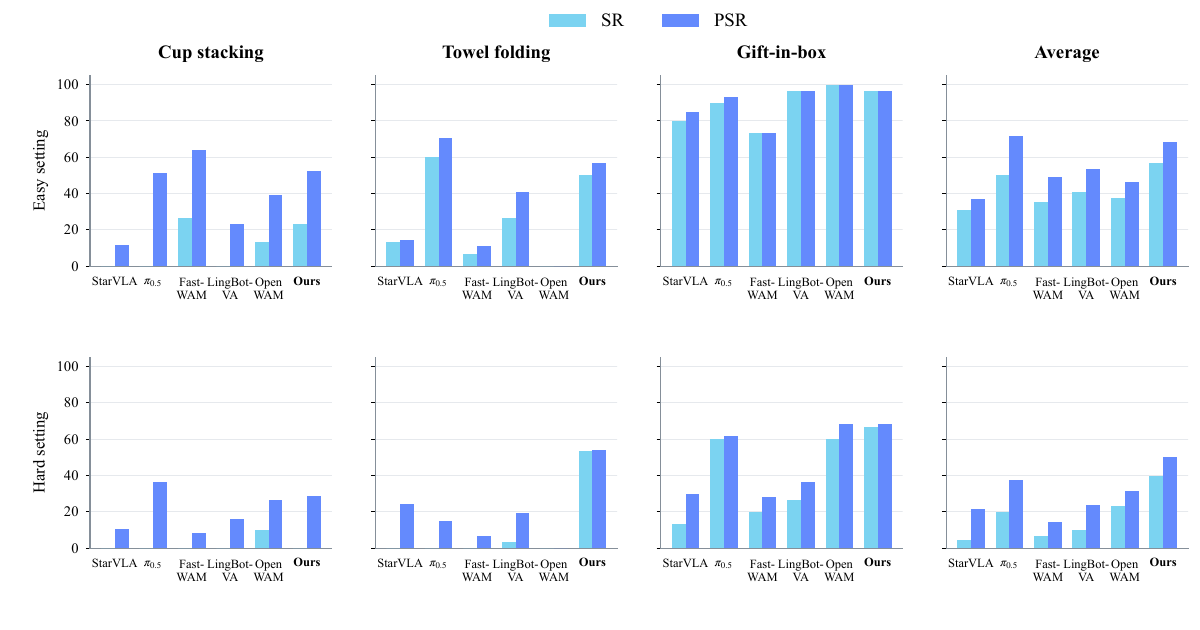}
    \caption{\textbf{Real-world comparison of our model and baseline methods.}}
    \label{fig:real-baselines-detailed}
\end{figure}
\FloatBarrier

\subsubsection{Pretraining Scaling}
\label{app:real-checkpoints}
Figure~\ref{fig:real-scaling-detailed} compares the 0-hour setting with models pretrained for 25k, 50k, 75k, and 100k steps, corresponding to approximately 22.5k, 45k, 67.5k, and 90k training video hours, respectively. At 0 hours, SR and PSR are both 0\% for every task under both settings.

\begin{figure}[!htbp]
    \centering
    \captionsetup{skip=5pt}
    \includegraphics[width=\textwidth,trim=0 4bp 0 0,clip]{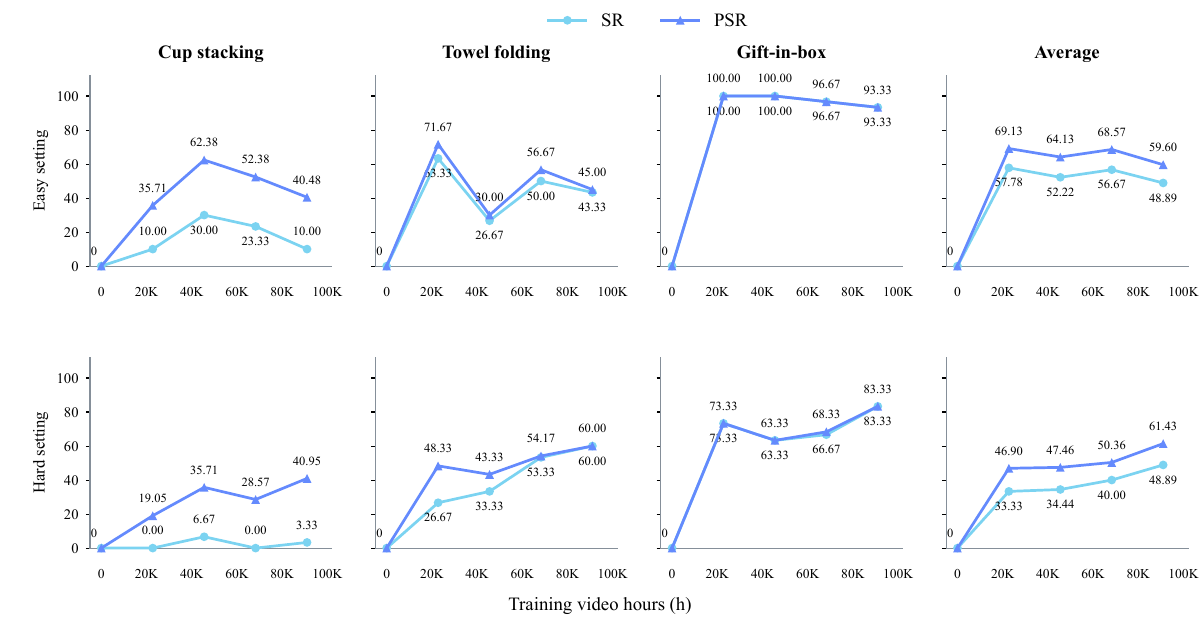}
    \caption{\textbf{Real-world performance with increasing pretraining video hours.}}
    \label{fig:real-scaling-detailed}
\end{figure}

\clearpage
\section{Real-World Execution Examples}
\label{app:real-execution-sequences}

Figures~\ref{fig:episode-cups}--\ref{fig:episode-gift} show execution examples under Hard settings.
Head-camera frames are sampled at uniform time intervals. 
Numbers indicate the order of the sampled frames.

\begingroup
\captionsetup[figure]{skip=2pt}
\noindent\begin{minipage}{\linewidth}
\centering
\includegraphics[width=.95\linewidth]{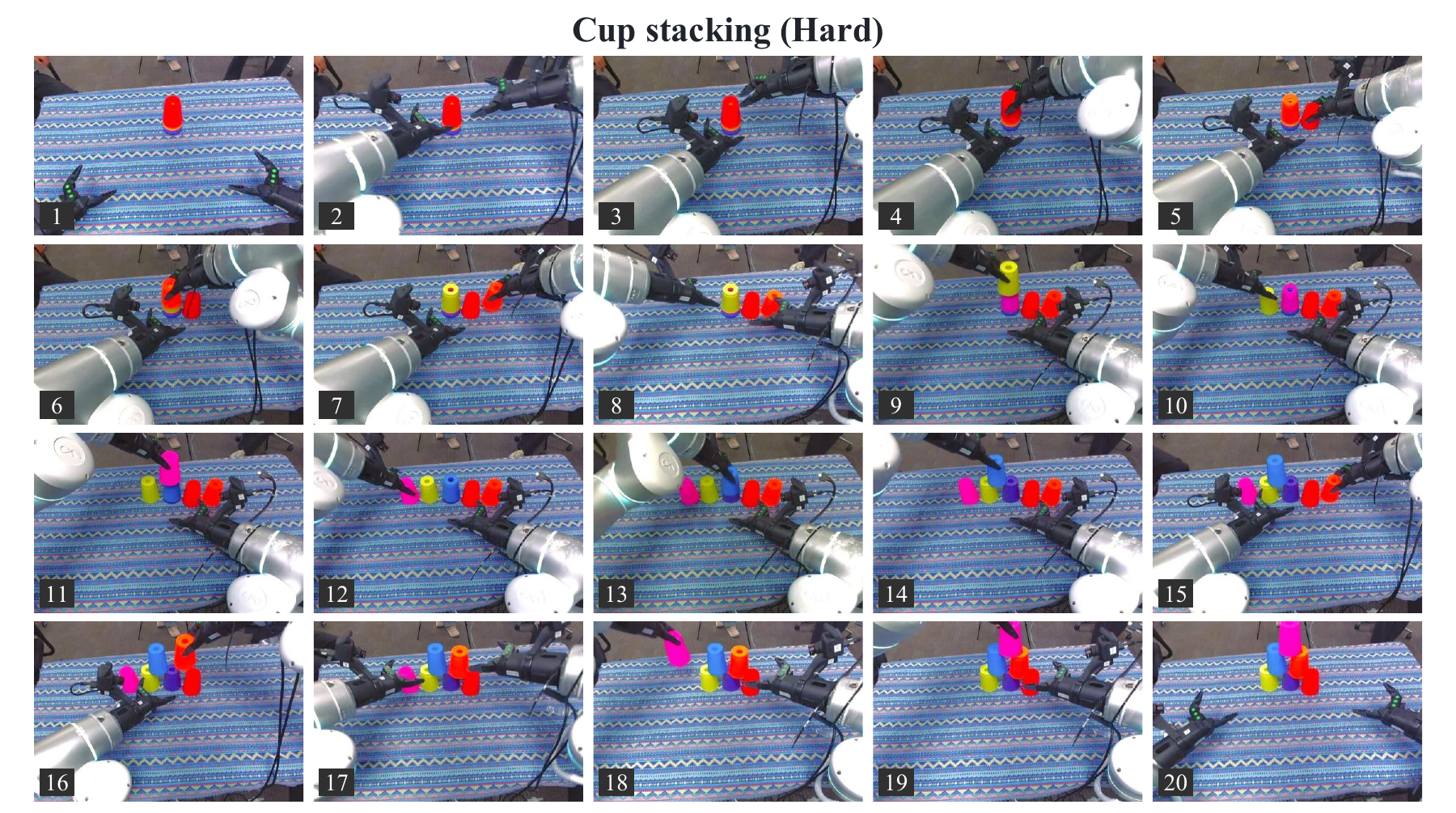}
\captionof{figure}{\textbf{Cup stacking.} The robot builds a three-tier pyramid from nested cups.}
\label{fig:episode-cups}
\end{minipage}\par\smallskip

\noindent\begin{minipage}{\linewidth}
\centering
\includegraphics[width=.95\linewidth]{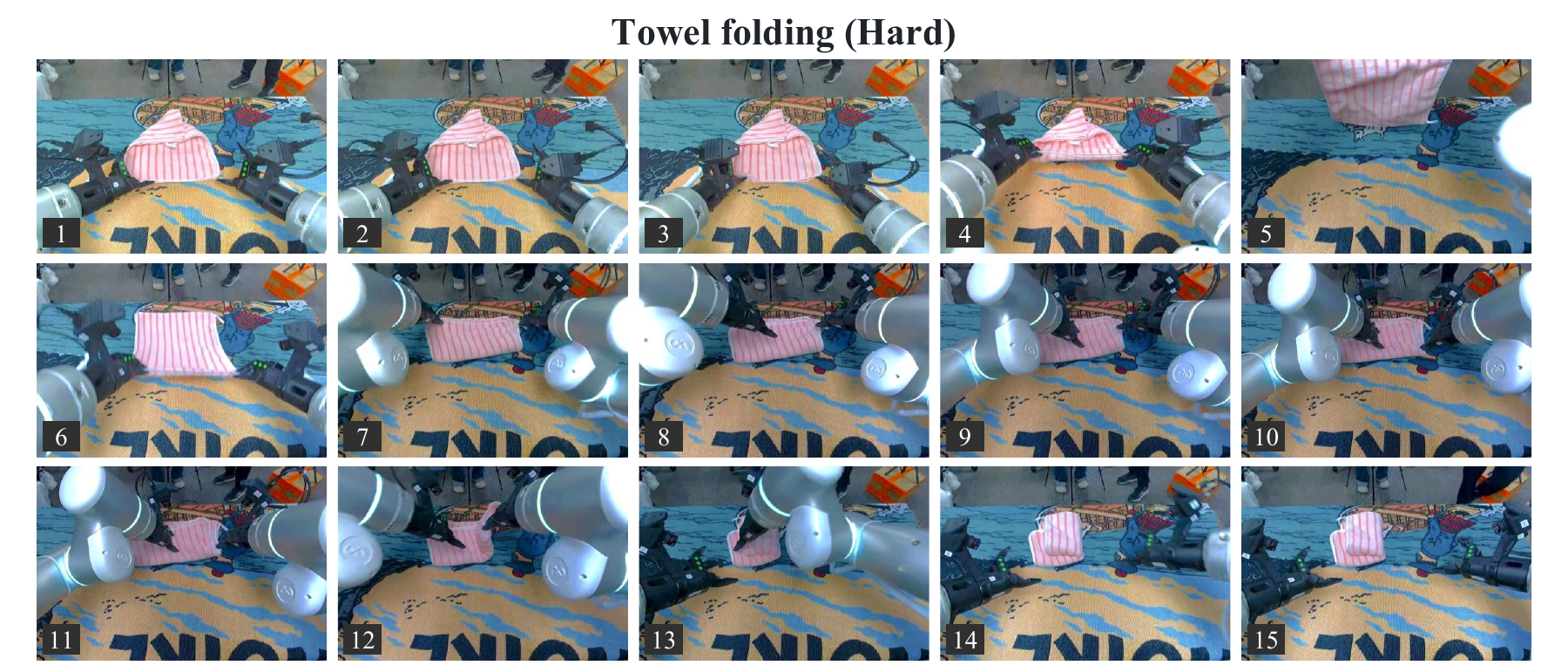}
\captionof{figure}{\textbf{Towel folding.} The robot spreads the towel and folds it twice.}
\label{fig:episode-towel}
\end{minipage}\par\smallskip

\noindent\begin{minipage}{\linewidth}
\centering
\includegraphics[width=.95\linewidth]{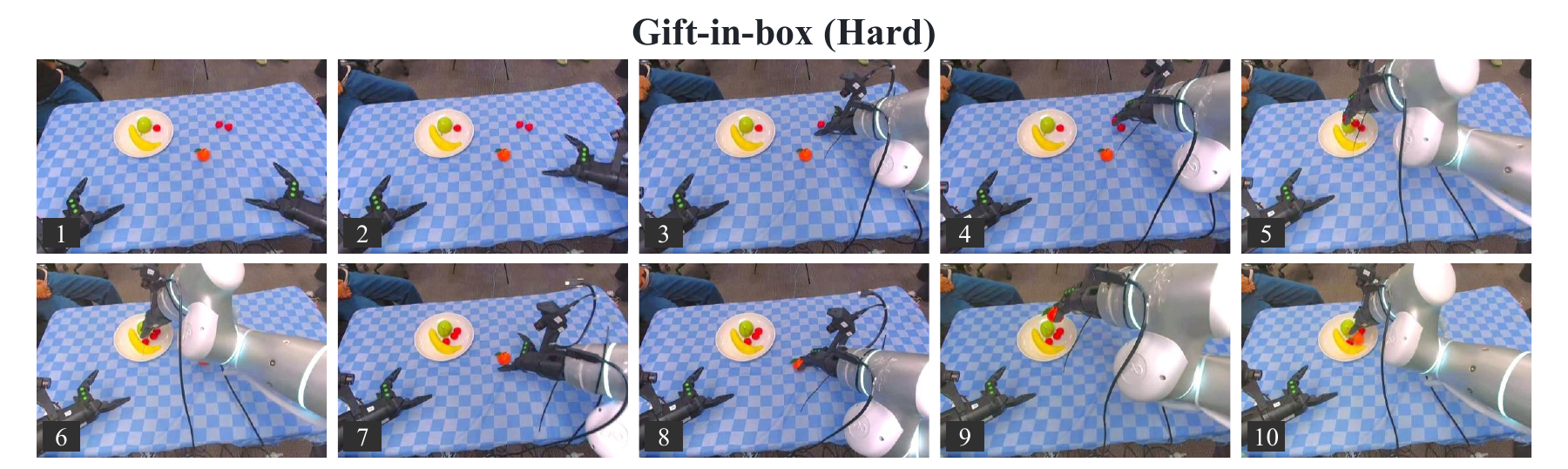}
\captionof{figure}{\textbf{Gift-in-box.} The robot places the gift in the box and releases it.}
\label{fig:episode-gift}
\end{minipage}\par
\endgroup

\endgroup

\end{document}